\documentclass[10pt,twocolumn,letterpaper]{article}
\usepackage[pagenumbers]{cvpr}
\usepackage[T1]{fontenc}
\usepackage{amsmath,amssymb,graphicx,booktabs,tabularx}
\usepackage{microtype}
\usepackage{dblfloatfix}
\newcounter{prop}
\newcounter{def}
\definecolor{cvprblue}{rgb}{0.21,0.49,0.74}
\usepackage[pagebackref,breaklinks,colorlinks,allcolors=cvprblue]{hyperref}
\def\confName{CVPR}
\def\confYear{2027}
\title{Attribution Through the Decoder:\\Prediction-Space Explanations for Structured Outputs}
\author{Zaer Bin Zaqy \qquad Brinnae Bent\\
Duke University\\
Durham, NC}

\hypersetup{pdfauthor={Zaer Bin Zaqy and Brinnae Bent}}
\begin{document}
\maketitle
\begin{abstract}
Networks that predict angles, gaze directions, and rotations do not output these quantities directly: they emit raw coordinates that a fixed decoder, such as normalisation, maps to the final prediction. Attribution methods are usually applied to the raw coordinates, so they can credit inputs for changes the decoder discards, such as rescaling a vector when only its direction matters. The resulting explanation can highlight inputs that have no effect on the prediction. We instead attribute through the decoder differential, which for normalisation decoders reduces to a scaled tangent projection. We characterise invariance to output rescaling, show that every fixed smooth scalar target on a sphere has points of zero tangential sensitivity, and show that an endpoint-projected variant of Integrated Gradients can still attribute to scale-only inputs, whereas direct angular Integrated Gradients cannot. On synthetic tasks with analytically known relevance, ambient attribution assigns up to 72\% of its mass to an irrelevant region. On MPIIGaze validation data, tangent Grad-CAM produces 44\% greater prediction displacement under 10\% pixel deletion. On PASCAL3D+ with a 6D rotation decoder, the full differential improves mean deletion displacement by $14.1^\circ$ over the ambient coordinate sum and by $13.8^\circ$ over a decoder-invariant scalar. Results for a two-polarity tangent LayerCAM variant are mixed: it exceeds ambient-sum LayerCAM on circular regression (one checkpoint for each of four backbones), but no difference is detected on quaternion pose. These results separate two failure sources, radial contamination and scalarisation loss. The approach uses existing attribution methods with decoder-aware targets, although some variants also require changes to attribution aggregation. An invariant attribution can still miss inputs that drive the prediction, so invariance and deletion performance should each be checked.
\end{abstract}

\section{Introduction}\label{sec:1}

Attribution methods explain a model's prediction by scoring how much each input feature, such as each pixel of an image, contributed to it. Grad-CAM \cite{selvaraju2017}, LIME \cite{ribeiro2016}, and Integrated Gradients (IG) \cite{sundararajan2017} are widely used examples, and practitioners rely on the resulting heatmaps to debug models and detect spurious features.

Each method explains a single scalar target, which in classification is naturally the predicted class score. For networks that predict angles, gaze directions, or 3D rotations, the usual target is formed from raw output coordinates, and we show that this can highlight inputs with provably no effect on the prediction.

The cause is a gap between what the network emits and what it predicts, and it produces two problems. First, a network layer can only emit a vector in $\mathbb{R}^n$, so the designer adds a fixed, parameter-free decoder: circular regression reads an angle from a plane vector, gaze and surface-normal models normalise a vector, and pose models project four, six, or nine numbers onto a rotation. The emitted coordinates have more degrees of freedom than the prediction, so many of them denote the same output: $z$ and $2z$ name the same direction. Attribution methods are nevertheless applied to the coordinates. An input that only changes the length of a gaze vector then receives credit, even though the decoder discards length entirely. Second, even when the network normalises its own output so that the length carries no sensitivity, a fixed scalar such as the coordinate sum registers change in the prediction along only one direction, and inputs that move the prediction in other directions go under-credited.

\begin{figure*}[tp]
\centering
\includegraphics[width=\linewidth]{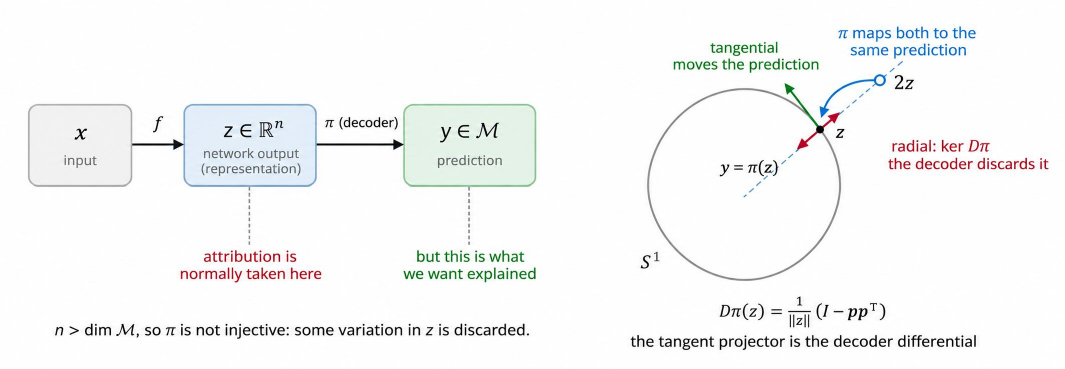}
\caption{Left: a structured-output model factors as $x \mapsto  z \mapsto  \pi (z)$; attribution is normally taken with respect to $z$, but the quantity to be explained is the decoded prediction $y$. Right: on $S^{1}$, $z$ and $2z$ denote the same prediction, so variation along the ray is discarded by the decoder; tangential variation moves the prediction, radial variation spans $\ker  D\pi $ and cannot. By the normalisation Jacobian (\S\ref{sec:2.1}) the tangent projector is the decoder differential up to a positive scalar.}
\label{fig:1}
\end{figure*}

Formally, we write the model as $x\mapsto z=f(x)\mapsto y=\pi(z)$, with $f$ the network, $z\in\mathbb{R}^{n}$ its ambient representation, $\pi$ the decoder, and $y\in M$ the structured prediction. The directions the decoder discards are $\ker D\pi(z)$: moving $z$ along them leaves $y$ unchanged to first order. An explanation of $y$ should ignore them, a requirement we call \emph{decoder invariance}, and it does so when attribution passes through the decoder differential, $D(\pi\circ f)=D\pi\cdot Df$. For a normalisation decoder $\pi(z)=z/\|z\|$, this amounts to projecting onto the tangent of the output sphere (Figure~\ref{fig:1}). We use \emph{prediction-space} as the umbrella term for methods that attribute through the decoder, and \emph{full differential} for the target $D(\pi\circ f)$. For normalisation decoders, \emph{tangent} describes its form and serves as the prefix for method names, such as tangent Grad-CAM. The same issue is latent in classification, where adding a constant to every logit leaves the softmax unchanged; structured outputs make it explicit.

Applied to a coordinate or a fixed combination of coordinates, none of these methods is decoder-invariant. Existing attribution methods can use decoder-aware targets, but retaining the full differential can require multiple backward passes and vector-valued aggregation; the two-polarity LayerCAM variant also changes polarity aggregation. We evaluate these adaptations, and we test whether decoder invariance alone is enough for an attribution to find the inputs whose removal changes the prediction.

Our contributions are:

\begin{enumerate}

\item We formulate attribution through the decoder and prove three results for normalisation decoders: invariance to output rescaling requires tangent target gradients (Proposition~\ref{prop:1}); any single scalar summary of the prediction has blind points where it registers no change (Proposition~\ref{prop:2}); and projecting only at the endpoint of Integrated Gradients still credits scale-only inputs, whereas integrating the decoder differential does not (Proposition~\ref{prop:3}).

\item We show that on synthetic tasks where the irrelevant inputs are known by construction, standard attribution places up to 72\% of its mass on them (\S\ref{sec:5}).

\item We identify when this failure occurs: across 35 controlled runs, it shrinks as the discarded and prediction-relevant pathways share more of their representation ($t(4)=-7.49$, $p=0.0017$; \S\ref{sec:6}).

\item We evaluate on gaze, circular regression, and 3D pose (\S\ref{sec:7}). Attributing through the decoder increases deletion displacement by 44\% on MPIIGaze validation participant p13 at 10\% deletion and by $14.1^\circ$ on 6D pose, while results on circular regression are mixed. An invariant scalar target recovers almost none of the pose gain ($13.8^\circ$ behind the full differential), so ignoring what the decoder discards is not enough, and invariance and deletion performance should each be checked.

\end{enumerate}

\subsection{Related Work}\label{sec:1.1}

\paragraph{Attribution.} Standard attribution methods explain a scalar target derived from the network output, whether by back-propagation (Grad-CAM \cite{selvaraju2017}, HiResCAM \cite{draelos2021}, LayerCAM \cite{jiang2021}), surrogate fitting (LIME \cite{ribeiro2016}, SHAP \cite{lundberg2017}), path integration (IG \cite{sundararajan2017}), or perturbation (occlusion \cite{zeiler2014}, RISE \cite{petsiuk2018}). Work on regression keeps the scalar-output assumption \cite{letzgus2022}. Govi et al. \cite{govi2023} adapt Grad-CAM to rotation regression and report difficulty selecting feature maps, which we read as an instance of the decoder-null issue. Manifold IG \cite{zaher2024}, Riemannian IG \cite{costanza2025} and least-resistance paths \cite{salek2025} make IG geometry-aware on the input side; we place the manifold on the output side, and the two compose.

\paragraph{Evaluation and Ground Truth.} Our real-task evaluation uses deletion \cite{petsiuk2018}. Because such proxies are known to be fragile \cite{hooker2019,adebayo2018,bilodeau2024}, we also construct ground truth. BAM \cite{yang2019} and Zhou et al. \cite{zhou2022} induce it as a property of the \emph{dataset}, and must argue that a trained model does not reach its accuracy by another route, an argument Zhou et al.~\cite{zhou2022} show can fail. Our irrelevant factor is a property of the \emph{output geometry}: the decoder discards the radial direction, so $\partial\hat\theta/\partial(\text{region B}) = 0$ identically for every model of this form.
\paragraph{Output Geometry.} Zhou et al. \cite{zhou2019} show that rotation representations in $\mathbb{R}^{d}$ for $d<5$ are discontinuous and propose the 6D alternative we use, and Levinson et al. \cite{levinson2020} analyse projection onto $\operatorname{SO}(3)$. Chen et al. \cite{chen2022} observe that training through such a projection shrinks the raw output norm unless regularised, which is the decoder-null freedom of \S\ref{sec:2} seen at training time. This literature studies representations and their optimisation; we take the decoder as given and study its consequences for attribution.

\section{Attribution Through the Decoder}\label{sec:2}

\subsection{Representation and Prediction}\label{sec:2.1}

Write a structured-output model as $x\mapsto z=f(x)\mapsto y=\pi(z)\in M$, with $f$ the network and $\pi$ a fixed decoder. We assume $\pi$ is smooth and a submersion on an open set $U\subseteq\mathbb R^n$ containing the image of $f$, with $n>\dim M$. The decoders considered here satisfy this assumption on their natural domains. The representation need not have unit norm.

\refstepcounter{def}\label{def:1}
\textbf{Definition~\ref{def:1}} (Decoder-null direction). A direction $\delta\in\mathbb R^n$ at $z$ is decoder-null if $\delta\in\ker D\pi(z)$.

Such a direction leaves the prediction fixed to first order. For normalisation, $\pi(z)=z/\|z\|$ and $p=z/\|z\|$, with $z\ne0$,
\begin{equation}
\begin{aligned}
D\pi(z)&=\frac{1}{\|z\|}(I-pp^\top)=\frac{1}{\|z\|}P_p,\\
\ker D\pi(z)&=\operatorname{span}\{p\}.
\end{aligned}
\end{equation}
Thus the decoder retains tangential sensitivity and discards radial sensitivity. Along a positive rescaling of $z$, the prediction is exactly unchanged. The full differential $D(\pi\circ f)$ depends only on the prediction map; the ambient derivative $Df$ can depend on its representation (Appendix~\ref{sec:N.1}).

\refstepcounter{prop}\label{prop:1}
\textbf{Proposition~\ref{prop:1}} (Gauge dependence). \emph{Let $\pi$ be the normalisation or circle-valued angle decoder, $z=f(x_0)\ne0$, and $\widetilde f=cf$, where $c>0$ is smooth and $c(x_0)=1$. Both networks give the same decoded prediction. For a differentiable scalar target $s$,}
\begin{equation}
\begin{aligned}
\nabla_x(s\circ\widetilde f)(x_0)={}&\nabla_x(s\circ f)(x_0)\\
&+\langle z,\nabla s(z)\rangle\nabla c(x_0).
\end{aligned}
\end{equation}
\emph{The input gradient is invariant to every such rescaling if and only if $\nabla s(z)\perp z$. Every target $s=h\circ\pi$ satisfies this condition.}

\refstepcounter{prop}\label{prop:2}
\textbf{Proposition~\ref{prop:2}} (Blind points). \emph{For smooth $s:\mathbb R^n\setminus\{0\}\to\mathbb R$, $n\ge2$, and every $r>0$, there are at least two points with $\|z\|=r$ at which the tangential component of $\nabla s(z)$ vanishes. At such a point the gradient is radial or zero. Nonzero linear targets $s(z)=v^\top z$ attain the minimum of two critical points, at $z=\pm rv/\|v\|$.}

Decoder invariance therefore does not prevent a fixed scalar target from losing tangential sensitivity. The full differential avoids selecting one fixed scalar direction, although subsequent CAM aggregation can still cancel contributions. Full statements, proofs, the gauge-fraction diagnostic, and the finite-perturbation remainder bound are in Appendix~\ref{sec:N.2}. For the local $S^1$ CAM below, tangent projection differs from the decoded-angle gradient by a positive per-image factor, preserving sign and rankings; this argument does not extend to integration or surrogate fitting across inputs with different output norms.

\subsection{\texorpdfstring{The $S^{1}$ Case and Three Failure Modes}{The S1 case and three failure modes}}\label{sec:2.2}

A circular regression model encodes the target as $(\sin  \theta , \cos  \theta )$, trains $f$: $\mathbb{R} ^{d} \to  \mathbb{R} ^{2}$ and recovers $\hat{\theta } = \operatorname{atan2}(\hat{y}_{\sin }, \hat{y}_{\cos })$. The raw output $z$ need not lie on $S^{1}$ (a tanh head bounds each coordinate but does not fix the norm), so write $p = z/\| z\| $. The unit tangent at $p$ is $t = (p_{\cos }, - p_{\sin })$ and the radial direction is $p$, so any $v \in  \mathbb{R} ^{2}$ decomposes as $v = \langle v,t\rangle t + \langle v,p\rangle p$ and only the tangential part affects the predicted angle (Figures~\ref{fig:1} and~\ref{fig:7}). The ratio of discarded to retained dimensions, $(n- m)/m$, is 1 on $S^{1}$, $\tfrac12 $ on $S^{2}$ and $\tfrac13 $ on $S^{3}$; it is a property of the parameterisation, and \S\ref{sec:7} shows it does \emph{not} predict empirical benefit.

\textbf{Failure mode 1: radial contamination and scalarisation loss.} Ambient attribution has two failure sources, and we use a separate baseline to expose each.

In the synthetic oracle (\S\ref{sec:5}, \S\ref{sec:6}) the baseline is \emph{ambient-gradient-magnitude}: the CAM weights each activation by $w = \sqrt{g_{\sin}^{2} + g_{\cos}^{2}}$, where $g = \partial z/\partial A$. This norm adds the squared tangential and radial components of $g$, so the map mixes sensitivity that moves the prediction with sensitivity the decoder discards. This is radial contamination. (A target equal to the output norm itself coincides with the radial control, Appendix~\ref{sec:L}.)

On the real tasks (\S\ref{sec:7}), and in the blind-point analysis, the baseline is \emph{ambient-sum}: Grad-CAM back-propagates the sum of the output coordinates, which on $S^{1}$ is $s(z) = z_{\sin} + z_{\cos}$. Its gradient is the fixed vector $v = (1, 1)$, so how much of it is tangential depends on where the prediction sits. On hard-normalised heads the radial Jacobian vanishes to machine precision, so radial contamination is absent and what remains is scalarisation loss. The prediction can move in $m$ tangent directions, but a fixed $v$ sees only the one along $P_{p} v$. That component has magnitude $\|v\| \sin \angle(p, v)$, and it vanishes at $p = \pm v/\|v\|$, where the target registers no change at all. This is Proposition~\ref{prop:2} in coordinates; Appendix~\ref{sec:6.4} and \S\ref{sec:7.3} confirm it on the oracle and on real pose data.

\textbf{Failure mode 2: Euclidean surrogate wrapping.} A scalar-angle LIME surrogate fits a linear model $\ell(b)=\beta_0+\beta^\top b$ to decoded angles, where $b$ indicates which superpixels are kept. Angles wrap around, so two nearly identical directions on either side of the branch cut have angle values almost $2\pi$ apart. A Euclidean regression treats this jump as a large change in the prediction and fits it. A KernelSHAP surrogate fitted to the principal angle has the same problem.

\textbf{Failure mode 3: metric breakdown.} Wrap-around also breaks naive metrics. Measured with $|\hat{\theta}_{1} - \hat{\theta}_{2}|$, a prediction that drifts from 23:50 to 00:10 appears to have moved 23.67 hours instead of 20 minutes. We therefore measure every displacement with the geodesic distance: on $S^{1}$, $\arccos(\sin\theta_{1} \sin\theta_{2} + \cos\theta_{1} \cos\theta_{2})$; on $S^{2}$, the angle between unit vectors; on $\operatorname{SO}(3)$, $2 \arccos |\langle q, q' \rangle|$. Angular error needs the same care, because taking the arccos of an unnormalised dot product lets the prediction norm leak into the angle (Appendix~\ref{sec:O.2}).

\section{Prediction-Space Attribution Methods}\label{sec:3}

We construct methods based on the decoder differential for local gradients, local surrogates, and path integration. Their invariance properties differ: raw-output circular surrogates are invariant only to a common rescaling across perturbations, and endpoint-projected IG is not decoder-invariant.

\subsection{Tangent-projected CAMs}\label{sec:3.1}

We contrast tangent-projected CAMs with the ambient-gradient-magnitude oracle baseline and ambient-sum real-task baseline defined in \S\ref{sec:2.2}. Applying the normalisation Jacobian (\S\ref{sec:2.1}) to a CAM, at $p$ with tangent $t$, we define

\begin{equation}
w^{\mathrm{tan}}_{kij}=\hat y_{\cos}\frac{\partial\hat y_{\sin}}{\partial A_{kij}}-\hat y_{\sin}\frac{\partial\hat y_{\cos}}{\partial A_{kij}}.
\end{equation}

Here $z = (\hat{y}_{\sin }, \hat{y}_{\cos })$ and $g = \partial z/\partial A$. For $z \ne  0$, $w^{\mathrm{tan}} = \| z\| \langle g,t\rangle  = \| z\| ^{2}\cdot \partial \hat{\theta }/\partial A$, so $w^{\mathrm{tan}}$ has no radial component and is proportional, with a positive per-image factor, to the gradient of the decoded angle; its sign indicates the direction of the local angular response to increasing the corresponding activation. We instantiate tangent Grad-CAM, HiResCAM, LayerCAM and saliency, and, as a control, radial variants that project onto $p$ instead of $t$. On $S^{2}$ and $S^{3}$ the map takes one backward pass per output coordinate, projects the $n$ per-location response fields with $P_{p} = I -  pp^{\top }$ at the decoded point and ranks by the norm of the projected vector. Applying LayerCAM's rectification directly would discard negative tangent contributions. The unrectified $S^{1}$ tangent LayerCAM entry is therefore the tangent HiResCAM map under another name (Table~\ref{tab:7}, Appendix~\ref{sec:H}). The two-polarity tangent LayerCAM variant keeps both polarities: it applies the rectification to $\operatorname{ReLU}(g)$ and $\operatorname{ReLU}(- g)$ of each tangent-coordinate gradient, forms a LayerCAM map from each and combines the half-maps by $L^{2}$ (two on $S^{1}$, six on the quaternion head, whose tangent basis is completed by Gram-Schmidt). It is invariant to $q \mapsto  - q$ and to signed permutations of the basis but not to general rotations of it (a $30{}^\circ $ rotation changes the map by about 11\% of its range), so we do not claim basis independence for it.

Tangent frames ($t$, $p$, or the Gram--Schmidt basis) are computed from the forward prediction and held fixed during differentiation. Projection is applied to per-coordinate response fields after CAM weighting; by linearity this is equivalent to projecting the gradients first. The ambient-sum LayerCAM control uses the standard one-polarity map $\operatorname{ReLU}(\sum_c\operatorname{ReLU}(g_c)\odot A_c)$. It shares per-location rectification with the two-polarity tangent variant but differs in polarity handling.

\subsection{Circular Surrogates}\label{sec:3.2}

Circular LIME fits two ridge regressions to the raw sine and cosine outputs over superpixel indicators. It decodes their angle with $\operatorname{atan2}$ and projects their coefficients at the unperturbed prediction:
$\phi_j=\hat y^{(0)}_{\cos}\beta_{\sin,j}-\hat y^{(0)}_{\sin}\beta_{\cos,j}$.
We use a von Mises locality kernel~\cite{fisher1993,mardia2000} on $S^1$; the spherical analogue regresses three coordinates and projects with $P_p$. Because the fits use raw outputs, the attribution rankings are invariant to a common positive rescaling across perturbations, but the attribution values are not. Input-dependent rescaling can also change the rankings. Appendix~\ref{sec:F.1} defines the spherical KernelSHAP baseline; the LIME kernel ablation is in Appendix~\ref{sec:F.2}.

\subsection{Angular Integrated Gradients}\label{sec:3.3}

Angular IG (AIG) has two forms. Direct AIG integrates $\partial\hat\theta/\partial x_i$ along $\gamma(\tau)=x'+\tau(x-x')$, with $\hat\theta$ continuously lifted along the path. Endpoint-projected (tangent) AIG instead combines raw-coordinate IG at the endpoint:
$\phi_i^{\mathrm{tan}}=\hat y_{\cos}\mathrm{IG}_{\sin,i}-\hat y_{\sin}\mathrm{IG}_{\cos,i}$.
On $S^1$ we use a blur integration baseline; deletion always sets selected pixels to zero. Baseline ablations are in Appendix~\ref{sec:E}.

\refstepcounter{prop}\label{prop:3}
\textbf{Proposition~\ref{prop:3}} (Direct angular IG). \emph{If $z(\tau)=f(\gamma(\tau))\ne0$ along the path,}
\begin{equation}
\begin{aligned}
\sum_i\phi_i^{\mathrm{dir}}&=\widetilde\theta(1)-\widetilde\theta(0),\\
\sum_i\phi_i^{\mathrm{tan}}&=rr'\sin(\hat\theta(x)-\hat\theta(x')),
\end{aligned}
\end{equation}
\emph{where $r,r'$ are the endpoint output norms. An input coordinate whose Jacobian column is decoder-null everywhere along the path receives zero direct attribution.}

Direct AIG accounts for lifted angular change, while endpoint projection depends on representation scale. For $f(a,b)=e^b(\sin a,\cos a)$ along the path from $(0,0)$ to $(\pi/2,1)$, endpoint projection assigns the scale-only input $b$ approximately $-2.56$; direct AIG assigns it zero. Appendix~\ref{sec:N.3} gives the full statement, proof, difference bound, and path diagnostics. This completeness guarantee is specific to $S^1$; the higher-dimensional IG baselines use the separately defined endpoint-projected targets (Appendix~\ref{sec:N.4}).

\section{Evaluation and Setup}\label{sec:4}

\begin{table*}[tp]
\centering
\caption{Oracle AMR, the fraction of absolute attribution mass on the prediction-relevant region (1.0 perfect, 0.5 chance, 0.0 all mass on the irrelevant factor). Rows group the three CAM/saliency variants, with ranges spanning the methods; the per-method values quoted in the text are Grad-CAM. SD $\le0.064$. The path-integral row accumulates tangent-Jacobian magnitudes at 32 path points; it is distinct from direct angular and endpoint-projected IG (Appendix~\ref{sec:N.4}).}
\label{tab:1}
\small
\setlength{\tabcolsep}{4pt}
\begin{tabularx}{\linewidth}{@{}Xccc@{}}
\toprule
Method & $S^{1} (n=500)$ & $S^{2} (n=30)$ & $\operatorname{SO}(3)$ via $S^{3} (n=200)$ \\
\midrule
Ambient-gradient-magnitude (Grad-CAM / HiResCAM / Saliency) & 0.428--0.520 & 0.281--0.508 & 0.747--0.757 \\
Tangent (all three) & \textbf{1.000} & \textbf{0.9996--0.9997} & \textbf{0.9998--0.9999} \\
Radial (all three) & 0.000 & 0.000--0.0003 & 0.0002--0.0005 \\
Circular / spherical LIME & 0.946 & 0.823 & 0.895 \\
Tangent-norm path integral & \textbf{1.000} & \textbf{0.9998} & $n/a$ \\
\bottomrule
\end{tabularx}
\end{table*}

We evaluate in two settings. On synthetic tasks we know by construction which pixels matter, so attribution can be scored against ground truth. On real tasks, ground-truth feature relevance is unavailable, so we measure how far the prediction moves when the highest-ranked pixels are removed.

\textbf{Metrics.} \emph{Oracle attribution mass ratio (AMR)} is the fraction of absolute attribution mass that falls on the region known to be relevant (1.0 is perfect, 0.5 is chance). \emph{Deletion displacement} measures how far the prediction moves in geodesic distance after setting a specified fraction of pixels, ranked by $|\phi_i|$, to zero in all channels in raw $[0,1]$ space. \emph{Mean deletion displacement} averages this quantity over the nine deletion fractions 10\%, 20\%, \ldots, 90\%. Higher is better: the attribution found pixels the prediction depends on. Units are hours on $S^1$ and degrees on $S^2$ and $\operatorname{SO}(3)$. \emph{Directional consistency} (DCS) tests whether the sign of an attribution is meaningful. On $S^1$ we delete each of the 100 most positively attributed pixels in turn and report the fraction for which the prediction moves against the attributed direction (chance 0.5). Changes are wrapped to $[-12,12)$ h; zero changes count as failures, and scores pool pixel deletions. Counts and the scoring formula are in Appendix~\ref{sec:Q}. On $S^2$ we instead report the cosine between the attributed tangent direction and the negative of the observed tangential displacement. \emph{Completeness} checks that path-integrated attributions sum to the change in the target each method is defined to explain. Implementation details are in Appendix~\ref{sec:Q}.

\textbf{Synthetic oracles.} Each $64\times64$ image has two halves. Region A encodes the direction to be predicted. Region B encodes only a magnitude, which sets the length of the output vector and is discarded by the decoder. The two halves pass through disjoint network branches, so region B provably cannot change the decoded prediction, and any attribution it receives is an error. We build one such task for each of $S^1$, $S^2$ and $\operatorname{SO}(3)$, the last with rotations represented as unit quaternions on $S^3$. Table~\ref{tab:1} uses $n=500/30/200$ test images respectively. The entangled controls, in which a single network sees both regions, use $n=200$ on $S^1$ and $S^2$, and a separately trained, wider-range stratified oracle supplies the $n=2{,}000$ alignment analysis (Appendix~\ref{sec:I.2}). Construction details are in Appendix~\ref{sec:L}.

\textbf{Plant ($S^1$).} The task is to predict the time of day from a plant image. The dataset has 3{,}765 masked plant images, split 2{,}621/767/377 into train, validation and test with each recording session kept in a single split. We train four backbones, each ending in a linear layer with two outputs followed by tanh. Attribution is computed for one checkpoint per backbone on the validation split, which was also used to select that checkpoint; accuracy is reported on the held-out test split. To delete a pixel we set it to zero in all channels, the background value of the pre-masked images; Appendix~\ref{sec:E} varies the IG integration baseline.

\textbf{Gaze ($S^2$).} On MPIIGaze~\cite{zhang2015} we train ResNet18 on one leave-one-person-out fold (participant p14 for test, p13 for validation), with five seeds for the Grad-CAM comparison. Accuracy is reported on test participant p14 ($n=3{,}000$). All gaze attribution metrics use validation participant p13 ($n=2{,}996$; also used for epoch selection).

\textbf{Pose ($\operatorname{SO}(3)$).} PASCAL3D+~\cite{xiang2014} provides 24{,}308 object crops from the nine categories without rotational symmetry. We split them 70/20/10 by source photograph (seed 42), so that crops from one photograph never appear in two splits, giving 4{,}886 validation crops. Quaternion heads are evaluated on ResNet18, ResNet50 and ViT-B/16, and a separate ResNet18-only 6D head uses Gram--Schmidt decoding~\cite{zhou2019}. Pose comparisons use five seeds. We evaluate rotation only, with the distance $2\arccos|\langle q,q'\rangle|$, which treats $q$ and $-q$ as the same rotation. Training settings and regression accuracies are in Appendices~\ref{sec:Q} and~\ref{sec:C}.
\section{When Ambient Attribution Fails}\label{sec:5}

\begin{table*}[tp]
\centering
\caption{Selected real-task deletion results in degrees across five seeds. Values are means, with seed SD where reported. The first row uses 10\% deletion; the remaining rows average the nine deletion fractions. Gaze rows use validation participant p13 (also used for epoch selection); gaze and quaternion rows use ResNet18 Grad-CAM; the 6D row compares attribution targets through the Gram-Schmidt decoder. The final column gives uncorrected paired seed-level $p$-values for the full differential against the ambient sum and, where evaluated, the decoder-invariant scalar. Detailed gaze and quaternion statistics in Appendix~\ref{sec:P}; the 6D comparison in Appendix~\ref{sec:N.6}.}
\label{tab:real-summary}
\small
\setlength{\tabcolsep}{4pt}
\begin{tabularx}{\linewidth}{@{}Xcccc@{}}
\toprule
Task / deletion & Ambient-sum & \shortstack{Decoder-invariant\\scalar} & \shortstack{Full\\differential} & \shortstack{$p$ vs. ambient\\/ invariant scalar} \\
\midrule
Gaze / 10\% & $4.52\pm0.36$ & -- & $6.50\pm0.29$ & $0.001$ / -- \\
Gaze / mean & $8.02\pm0.65$ & -- & $9.08\pm0.72$ & $0.017$ / -- \\
Quaternion pose / mean & $47.77\pm3.48$ & -- & $56.55\pm4.82$ & $0.013$ / -- \\
6D pose / mean & $48.9$ & $49.2$ & $63.0$ & $0.008$ / $<0.001$ \\
\bottomrule
\end{tabularx}
\end{table*}

In the synthetic oracle the two network branches see disjoint pixels, so the derivative of the decoded prediction with respect to every region-$B$ pixel is exactly zero. Any attribution placed on region $B$ is therefore an error, and a method that respects the decoder should score an AMR of 1.

Tangent projection scores at least 0.9996 on all three manifolds (Table~\ref{tab:1}), which confirms that the measurement works. Ambient attribution scores far lower. With Grad-CAM its AMR is 0.498 on $S^{1}$, no better than chance, and 0.281 on $S^{2}$, below chance: on $S^{2}$, 72\% of the attribution mass lies on pixels that cannot affect the prediction.

The radial control shows where that mass comes from. Projecting the gradient onto the radial direction, the one the decoder discards, gives $\mathrm{AMR} \le 0.0005$, so nearly all of its mass lands on region $B$. The ambient weight adds the squared tangential and radial components (\S\ref{sec:2.2}), so its map is a mixture of the two and its AMR falls between them. This is the radial contamination of failure mode 1, measured directly (Figure~\ref{fig:6}, Appendix~\ref{sec:A}).

Circular and spherical LIME score 0.823--0.946, most of the way to the tangent result. Ambient scores are higher on $S^{3}$ (0.747--0.757) than on the other two manifolds. The area occupied by region $A$ does not account for this and we have not identified the cause, so we report the differences across manifolds without interpreting them.
\section{What Determines the Failure}\label{sec:6}

\begin{figure*}[tp]
\centering
\includegraphics[width=\linewidth]{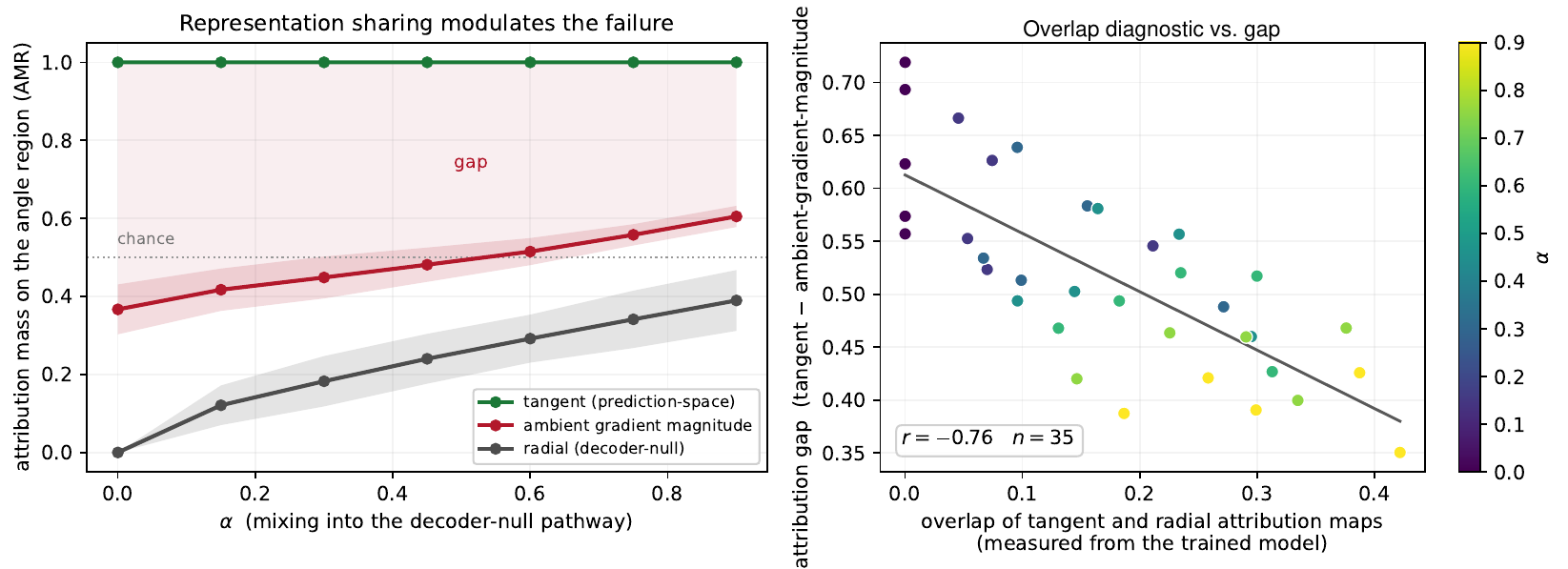}
\caption{Controlled representational separability. Left: attribution mass on the angle-encoding region against $\alpha $, the amount of prediction-relevant representation mixed into the decoder-null pathway (5 seeds per point, mean $\pm $ SD). Tangent is flat at 1.000 by construction; radial climbs off zero as the null pathway begins reading region-A features; ambient, a mixture, follows. The shaded band is the gap. Right: the same gap against the mean cosine similarity of tangent and radial saliency maps over the first 40 test images per run (pooled Pearson $r=-0.76$; definition in Appendix~\ref{sec:I.1}). All 35 runs are shown.}
\label{fig:2}
\end{figure*}

In \S\ref{sec:5} the prediction-relevant and decoder-null information sit in disjoint pixels and pass through separate branches. When a single network sees both regions instead (the entangled controls, Appendix~\ref{sec:D}), the failure almost disappears: the tangent, ambient, and radial variants of Grad-CAM and HiResCAM differ by at most 0.007 AMR within each method and manifold. Saliency differs more, by 0.067 on $S^{1}$ and 0.056 on $S^{2}$. The two architectures also differ in depth, pooling, and parameter count, however, so this comparison cannot show that sharing is the cause. We therefore vary the amount of sharing inside one architecture.

\subsection{A Controlled Interpolation}\label{sec:6.1}

The oracle computes a unit direction from region $A$, angle $= \operatorname{unit}(\operatorname{ahead}(a_{A}))$, and a positive magnitude from region $B$, $\operatorname{mag} = \operatorname{softplus}(\operatorname{mhead}(a_{B}))$, and emits $z = \operatorname{mag}\cdot$angle. Because $\operatorname{mag}$ scales $z$ without rotating it, the magnitude branch is exactly the direction the decoder discards. To introduce sharing, we mix region-$A$ activations into the input of the magnitude head, $a_{B}^{\alpha} = (1 - \alpha) a_{B} + \alpha a_{A}$, for $\alpha$ from 0 to 0.9, and leave the angle path untouched. (At $\alpha = 1$ the magnitude pathway would be replaced, not shared.)

Mixing in this one direction keeps the ground truth intact. The angle head never sees region $B$, so $\partial\hat{\theta}/\partial B = 0$ at every $\alpha$ (the largest ratio to $\partial\hat{\theta}/\partial A$ across all 35 runs is $2.0\times 10^{-7}$), and AMR means the same thing throughout. Mixing in both directions would not preserve this: at $\alpha = 0.9$ region $B$ would influence the prediction $10.8\times$ more strongly than region $A$, and attribution on $B$ would be correct.

\subsection{Representation Sharing Modulates the Failure}\label{sec:6.2}

We train 5 seeds at each of 7 values of $\alpha$ (35 runs, all converged). For each run we compute the AMR gap, defined as tangent AMR minus ambient-gradient-magnitude AMR, using input-gradient saliency on 100 test images per run. The gap shrinks steadily as sharing increases, from $+0.633$ at $\alpha = 0$ to $+0.395$ at $\alpha = 0.9$ (Figure~\ref{fig:2}). The condition means are close to linear in $\alpha$ ($r = -0.997$, $p = 6.6\times 10^{-7}$), and the decline is monotone in 4 of 5 individual seeds. It is also significant at the seed level, both for the per-seed slope ($t(4) = -7.49$, $p = 0.0017$) and in a within-seed permutation test ($p = 5.0\times10^{-6}$; $k = 0$ of 200{,}000 relabellings; Appendix~\ref{sec:I}).
The $\alpha = 0$ gap is a mean over five saliency runs. It differs from the saliency entry in Table~\ref{tab:1} (ambient AMR 0.428, gap 0.572) because the trained models and test images differ. 

Tangent AMR is exactly 1.000 at every $\alpha$, as first-order invariance requires (\S\ref{sec:2.1}). Radial AMR rises from 0.000 to 0.390 as the magnitude head starts to read region-$A$ features, and ambient attribution, which mixes the tangent and radial components, rises with it. Model accuracy does not explain the trend: we detect no association between angular error and $\alpha$ ($r = -0.25$, $p = 0.15$). The effect also replicates with HiResCAM (gap $0.541 \to 0.238$).

Sharing reduces the failure without removing it. The gap falls by 38\% over the tested range, and the decline is still approximately linear at $\alpha = 0.9$, leaving the gap well above the near-zero level of the entangled controls. The sweep therefore shows that sharing modulates the failure.

Three diagnostics computed on the trained model track the gap within this sweep, but none is yet validated as a predictor on real architectures (Appendix~\ref{sec:6.3}). On quaternion pose, lower scalar-target alignment is associated with a larger deletion gap, but low alignment also coincides with harder poses, so the association does not isolate the cause (Appendix~\ref{sec:6.4}).

\section{Real Structured-Prediction Tasks}\label{sec:7}

We compare methods using deletion displacement (Table~\ref{tab:real-summary}). We distinguish attribution targets from spatial weighting, rectification, and polarity aggregation.

\subsection{\texorpdfstring{Real $S^2$: Gaze Estimation}{Real S2: Gaze Estimation}}\label{sec:7.1}

On validation participant p13 ($n=2{,}996$), across five ResNet18 seeds, tangent Grad-CAM produces 44\% greater displacement than ambient-sum Grad-CAM at 10\% deletion, in every seed ($p=0.001$). The advantage holds at the uncorrected $0.05$ level through fraction 0.5, and the sign reverses from 0.8 onward, though not significantly, so the gain is specific to early deletion. Because the head is hard-normalised, its radial derivative is zero, and the difference reflects tangential sensitivity lost through a fixed scalar readout. Perturbation baselines were evaluated on one checkpoint. A centred spherical-occlusion estimator scores higher than the original estimator (mean $11.02^\circ$ versus $9.37^\circ$) and exceeds tangent Grad-CAM at every fraction (Appendix~\ref{sec:P.1}, which includes seed-versus-image uncertainty analysis).

\subsection{\texorpdfstring{Real $S^1$: Deletion and Direction}{Real S1: Deletion and Direction}}\label{sec:7.2}

Table \ref{tab:3} reports one checkpoint
per backbone on the 767-image plant validation split, also used for
checkpoint selection. The two-polarity tangent LayerCAM gain changes both the target and polarity aggregation. Plain tangent Grad-CAM has mixed results. Circular LIME improves on scalar-angle LIME across these checkpoints, while IG has higher mean displacement than every CAM on all four backbones. Direct AIG completes to lifted angular change, but its unsigned mean displacement differs from the best reported standard scalar-target IG by at most 0.04 h. Directional reliability is separate: the DCS estimates of 0.52--0.53 are not distinguishable from chance under the approximate image-effective intervals (Appendix~\ref{sec:P.2}), and tangent-CAM direction varies across checkpoints. Full method comparisons and baseline ablations are in Appendix~\ref{sec:P.2}.

\begin{figure*}[t]
\centering
\includegraphics[width=\linewidth]{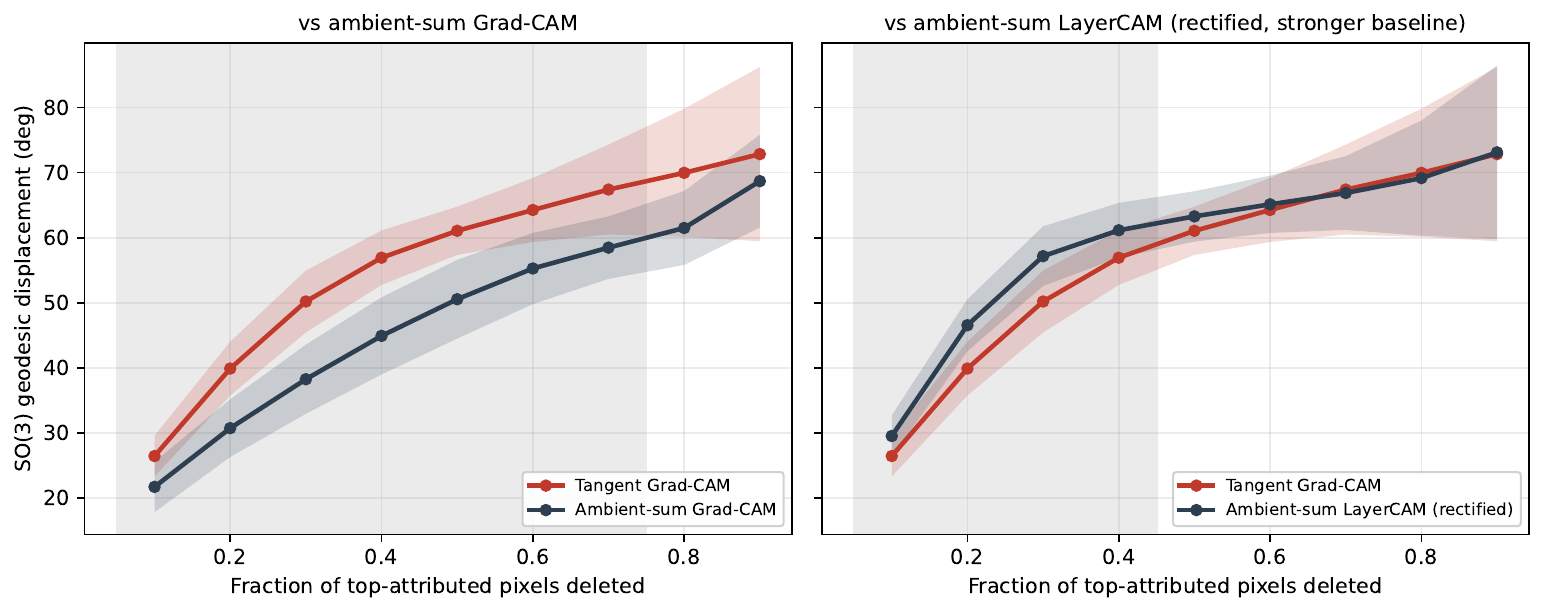}
\caption{PASCAL3D+ $\operatorname{SO}(3)$ deletion curves, five seeds (mean $\pm $ SD, $n = 4{,}886$), tangent Grad-CAM against the two ambient baselines. Left: against ambient-sum Grad-CAM, the mean tangent curve leads at every fraction; shading marks seed-level $p < 0.05$. Right: against ambient-sum LayerCAM, whose  weighting is the stronger baseline, the plain tangent map trails at low fractions (shaded, $p < 0.05$) while no difference is detected for the separate two-polarity tangent comparison (text).}
\label{fig:4}
\end{figure*}

\subsection{\texorpdfstring{Real $\operatorname{SO}(3)$: Object Pose}{Real SO(3): Object Pose}}\label{sec:7.3}

The 6D head lets us compare three targets through the same decoder. Mean deletion displacement is $63.0^\circ$ for the full differential, $48.9^\circ$ for the ambient coordinate sum, and $49.2^\circ$ for the decoder-invariant rotation-matrix sum (Table~\ref{tab:real-summary}). The full differential gains $14.1^\circ$ over the ambient sum and $13.8^\circ$ over the invariant scalar, and the two scalars do not differ detectably ($p=0.85$). Making a scalar target decoder-invariant does not recover what the full differential captures.
On the ResNet18 quaternion head, tangent Grad-CAM exceeds ambient-sum Grad-CAM in all five seeds at fractions 0.1--0.7 (paired $p<0.05$), with a mean gain of $8.78^\circ$ ($p=0.013$; Figure~\ref{fig:4}). Ambient-sum LayerCAM is a stronger baseline: it exceeds the plain tangent map through fraction 0.6, and no difference is detected between it and the two-polarity tangent variant (mean difference $-0.08^\circ$, $p=0.75$; 90\% CI $[-0.58^\circ,+0.42^\circ]$). That variant also changes polarity aggregation and is not invariant to rotations of its tangent basis. Results on other backbones are less clear: the ResNet50 comparison is inconclusive ($p=0.094$) and ViT results depend on the CAM construction. Single-checkpoint perturbation methods also remain competitive, so the gains are limited to the comparisons stated (Appendix~\ref{sec:P.3}).

\section{Discussion}\label{sec:9}

Attribution maps are used to debug models and detect spurious features, and for structured outputs the usual target can undermine both uses. A heatmap computed on raw coordinates can highlight inputs that cannot change the prediction, sending a practitioner after features the model does not use. On synthetic tasks where the irrelevant inputs are known by construction, the ambient methods we tested place 24--72\% of their attribution mass on such inputs (\S\ref{sec:5}). Across 35 controlled runs, this error shrinks as the two pathways share more of their representation (\S\ref{sec:6}). The approach requires no model retraining. It adapts existing attribution methods with decoder-aware targets, while some variants require multiple backward passes or changes to attribution aggregation. The real tasks reveal a second problem beyond radial contamination: scalarisation loss. The gaze and quaternion heads normalise their output, leaving no radial sensitivity to remove, yet a single scalar target still sees only one direction in which the prediction can move. On 6D pose the full differential beats both scalar targets, including the decoder-invariant one. Decoder invariance therefore says what an attribution means, not how well it performs under deletion, and the two should be checked separately. Our gains are limited to the comparisons reported. On quaternion pose the ResNet50 result is inconclusive, and the two-polarity tangent variant shows no detectable difference from ambient-sum LayerCAM. On $S^{1}$ that variant scores higher on every checkpoint, but it also changes polarity aggregation and is not seed-replicated. Direct AIG is complete for the lifted angle, but its unsigned mean displacement stays close to standard IG and the reliability of its sign varies.

\begin{table}[t]
\centering
\small
\setlength{\tabcolsep}{0pt}
\caption{Plant $S^1$ mean deletion displacement (hours; single checkpoint per backbone,
$n=767$ validation images also used for checkpoint selection). Each pair
gives a standard target, then its prediction-space counterpart. Full table with directional
consistency: Table~\ref{tab:7}.}
\label{tab:3}
\begin{tabular*}{\linewidth}{@{\extracolsep{\fill}}lcccc@{}}
\toprule
Method & R18 & R50 & EffV2-S & ViT-B \\
\midrule
Ambient-sum Grad-CAM       & 2.10 & 0.94 & 0.65 & 1.10 \\
Tangent Grad-CAM           & 1.58 & 1.32 & 1.94 & 0.73 \\
\addlinespace[2pt]
Ambient-sum LayerCAM       & 5.06 & 4.45 & 3.53 & 0.75 \\
Two-polarity tangent LayerCAM & 5.24 & 4.53 & 4.33 & 1.44 \\
\addlinespace[2pt]
Scalar-angle LIME          & 2.00 & 2.23 & 2.45 & 4.02 \\
Circular LIME              & 2.32 & 2.36 & 2.68 & 4.78 \\
\addlinespace[2pt]
Scalar-target IG           & 6.16 & 7.06 & 6.34 & 8.43 \\
AIG, direct         & 6.16 & 7.07 & 6.36 & 8.39 \\
\bottomrule
\end{tabular*}
\end{table}

Future work should extend the real-task evaluation beyond the three tasks and single dataset per task studied here, and replicate across seeds the $S^{1}$ attribution experiments and $S^{2}$ perturbation baselines. Extending tasks with analytically known relevance beyond synthetic images would allow the cases where deletion rankings and decoder invariance disagree (Appendix~\ref{sec:O.1}) to be resolved against ground truth.

\section*{Acknowledgments}

We thank Lucas Bauer and Colleen Doherty for their contributions to the data collection efforts that helped inspire and inform this project.

\label{main-text-end}\par
\clearpage
{\small
\bibliographystyle{ieeenat_fullname}
\bibliography{references}
}
\clearpage
\maketitlesupplementary
\label{supplement-start}
\appendix
\setcounter{table}{0}
\setcounter{figure}{0}
\renewcommand{\thetable}{S\arabic{table}}
\renewcommand{\thefigure}{S\arabic{figure}}
\renewcommand{\theHtable}{supp.\arabic{table}}
\renewcommand{\theHfigure}{supp.\arabic{figure}}
\section{Supplementary Figures and Tables}\label{sec:A}

Figures and tables referenced from the main text.

\begin{figure*}[tp]
\centering
\includegraphics[width=\linewidth]{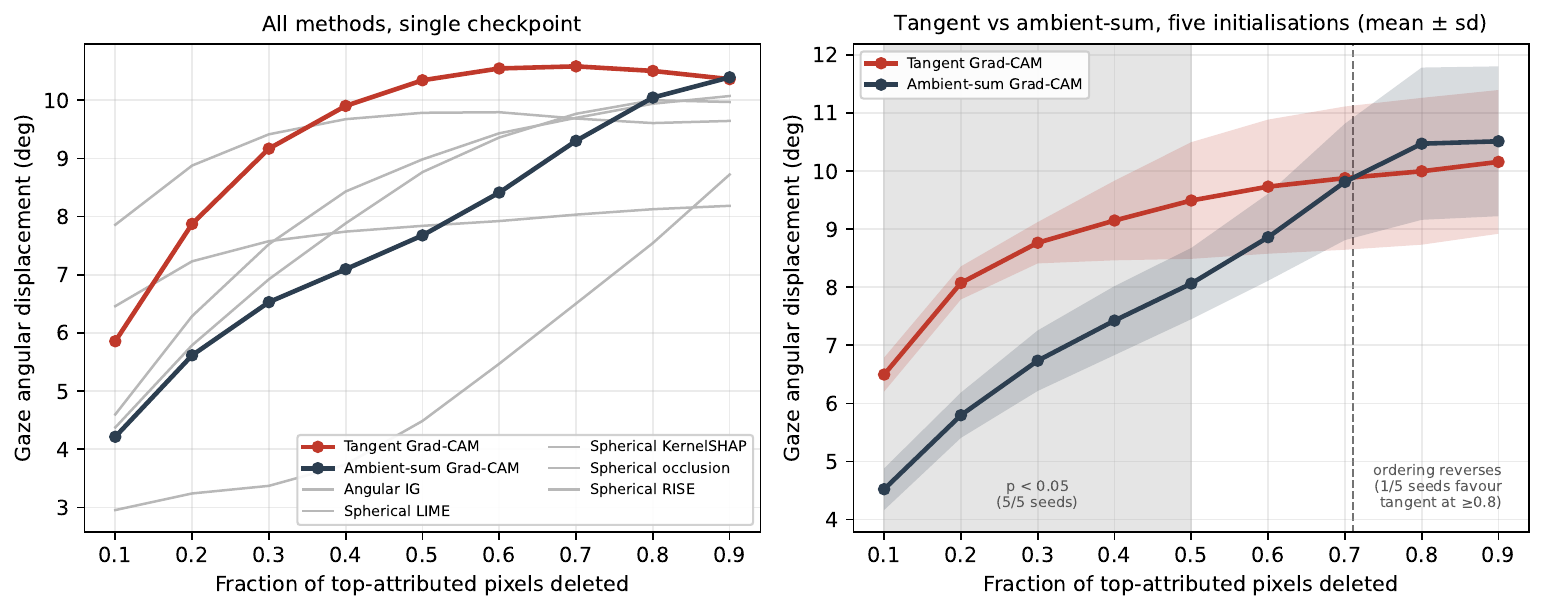}
\caption{MPIIGaze $S^{2}$ deletion on validation participant p13 ($n=2,996$ per run, also used for epoch selection), higher means greater deletion displacement. Left: all seven methods on the original unseeded checkpoint (not one of the five seeded models), the only run in which the perturbation baselines were scored, so the occlusion comparison of \S\ref{sec:7.1} rests on this panel alone. Right: tangent vs ambient-sum Grad-CAM over five independent initialisations sharing one LOPO split, mean $\pm $ SD across seeds. Shading marks fractions 0.1--0.5, where the paired difference reaches $p<0.05$ over seeds and favours tangent in $5/5$; the dashed line marks where the ordering reverses, with $1/5$ seeds favouring tangent beyond it. Per-fraction values and tests are in Table~\ref{tab:8}.}
\label{fig:5}
\end{figure*}

\begin{figure*}[tp]
\centering
\includegraphics[width=\linewidth]{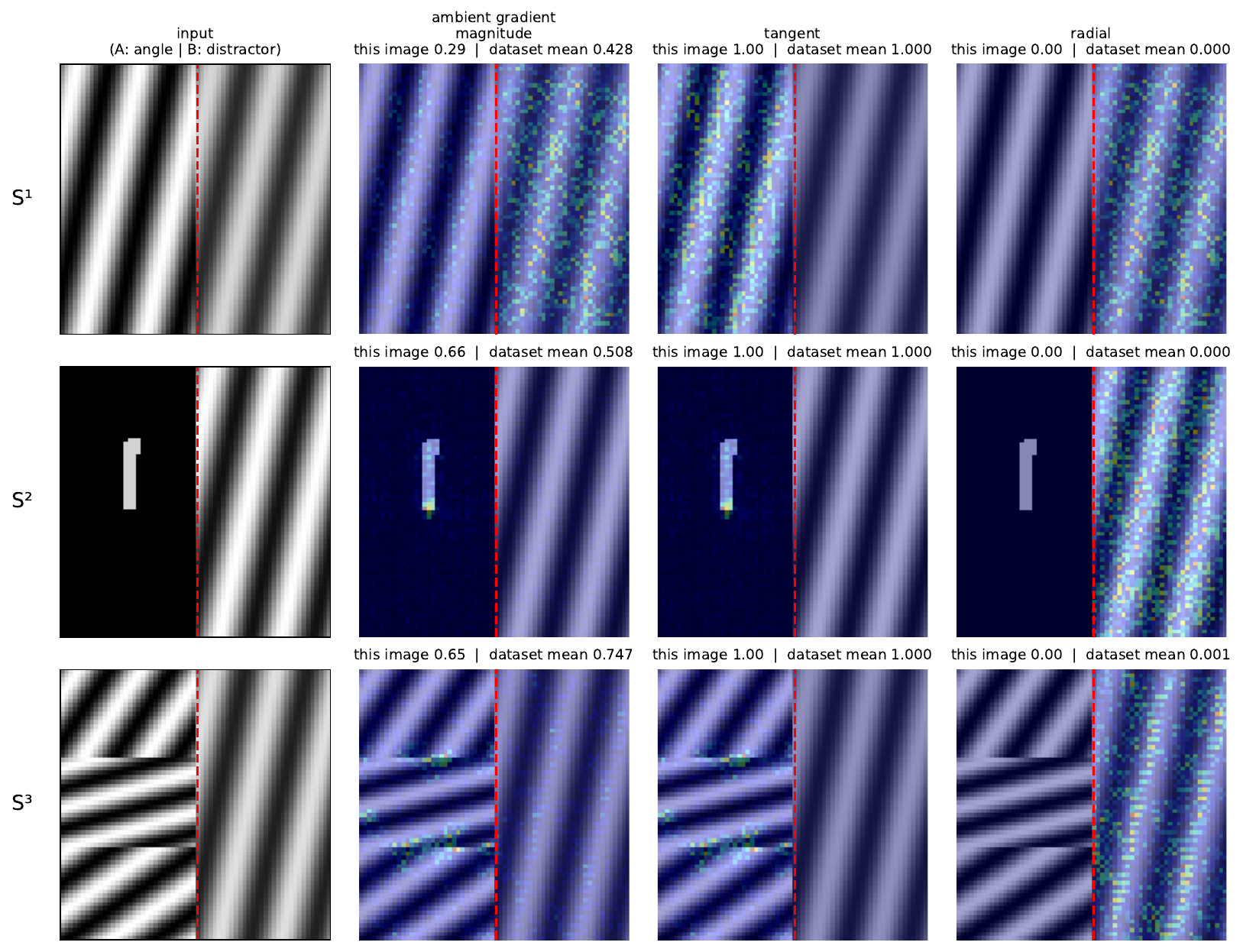}
\caption{Oracle attribution on all three manifolds (rows: $S^{1}$, $S^{2}$, $S^{3}$). Columns: input, ambient-gradient-magnitude saliency, tangent saliency, radial saliency. The dashed line marks the region boundary, region A (left) provably encodes the angle and region $B$ (right) provably encodes only the irrelevant magnitude factor. Radial attribution lands on region $B$ in every row (it recovers exactly the factor that cannot influence the prediction) while ambient-gradient-magnitude attribution spreads across both. Each panel is inset with the AMR for that single test image alongside the saliency dataset mean from the Table~\ref{tab:1} runs. The panel models are separate trainings (4,000 images, 12 epochs). The two can diverge substantially: the $S^{2}$ ambient-gradient-magnitude panel scores 0.66 on this image against a saliency dataset mean of 0.508. The panels illustrate individual examples; the table reports dataset means.}
\label{fig:6}
\end{figure*}

\begin{figure*}[tp]
\centering
\includegraphics[width=0.52\linewidth]{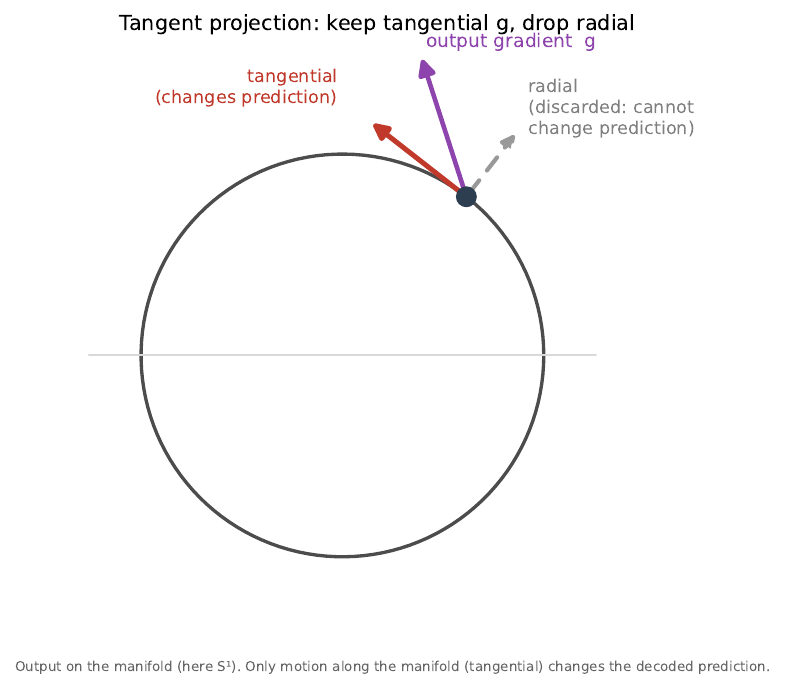}
\caption{Tangent--radial decomposition of an output gradient at a point $p$ on $S^{1}$. The tangential component $\langle  g,t\rangle  t$ is the only part that changes the predicted angle, the radial component $\langle  g,p\rangle  p$ changes the ambient representation without changing the decoded angle $\hat{\theta }$. Ambient-coordinate attribution combines both. Projecting onto $T_{p} S^{1}$ discards the radial part by construction.}
\label{fig:7}
\end{figure*}

\begin{table*}[tp]
\centering
\caption{MPIIGaze $S^{2}$ per-fraction deletion (degrees, higher means greater deletion displacement), ResNet18, $n=2,996$. Validation participant p13, also used for epoch selection. This is the original unseeded checkpoint (validation error $5.49^\circ$, test error $5.20^\circ$), not one of the five seeded models. The seed-replicated tangent/standard comparison, which supersedes the Mean column here, is Table~\ref{tab:8}. Bold marks the best value per fraction. The Mean column is left unbolded deliberately, as \S\ref{sec:7.1} argues averaging over the full range mixes opposite-signed regimes. The spherical occlusion row is the earlier estimator (patch 16, stride 8, bilinear upsampling), also shown in Figure~\ref{fig:5}; the centred estimator is reported below. Both use black fill and unsigned geodesic displacement, differing in grid resolution and patch-to-pixel mapping.}
\label{tab:5}
\small
\setlength{\tabcolsep}{4pt}
\begin{tabularx}{\linewidth}{@{}Xcccccccccc@{}}
\toprule
Method & 0.1 & 0.2 & 0.3 & 0.4 & 0.5 & 0.6 & 0.7 & 0.8 & 0.9 & Mean \\
\midrule
Tangent & 5.86 & 7.88 & 9.17 & \textbf{9.90} & \textbf{10.34} & \textbf{10.55} & \textbf{10.58} & \textbf{10.51} & 10.36 & 9.46 \\
Spherical occlusion & \textbf{7.86} & \textbf{8.88} & \textbf{9.42} & 9.68 & 9.79 & 9.80 & 9.69 & 9.61 & 9.65 & 9.37 \\
Spherical LIME & 4.60 & 6.29 & 7.53 & 8.43 & 8.98 & 9.43 & 9.70 & 9.94 & 10.07 & 8.33 \\
Spherical KernelSHAP & 4.38 & 5.79 & 6.92 & 7.88 & 8.77 & 9.36 & 9.77 & 10.00 & 9.97 & 8.09 \\
Ambient-sum & 4.21 & 5.62 & 6.53 & 7.10 & 7.68 & 8.41 & 9.30 & 10.05 & \textbf{10.39} & 7.70 \\
Endpoint-projected IG & 6.46 & 7.23 & 7.58 & 7.74 & 7.84 & 7.92 & 8.03 & 8.13 & 8.19 & 7.68 \\
Spherical RISE & 2.95 & 3.24 & 3.37 & 3.74 & 4.49 & 5.47 & 6.51 & 7.55 & 8.72 & 5.12 \\
\bottomrule
\end{tabularx}
\end{table*}

\begin{figure*}[tp]
\centering
\includegraphics[width=\linewidth]{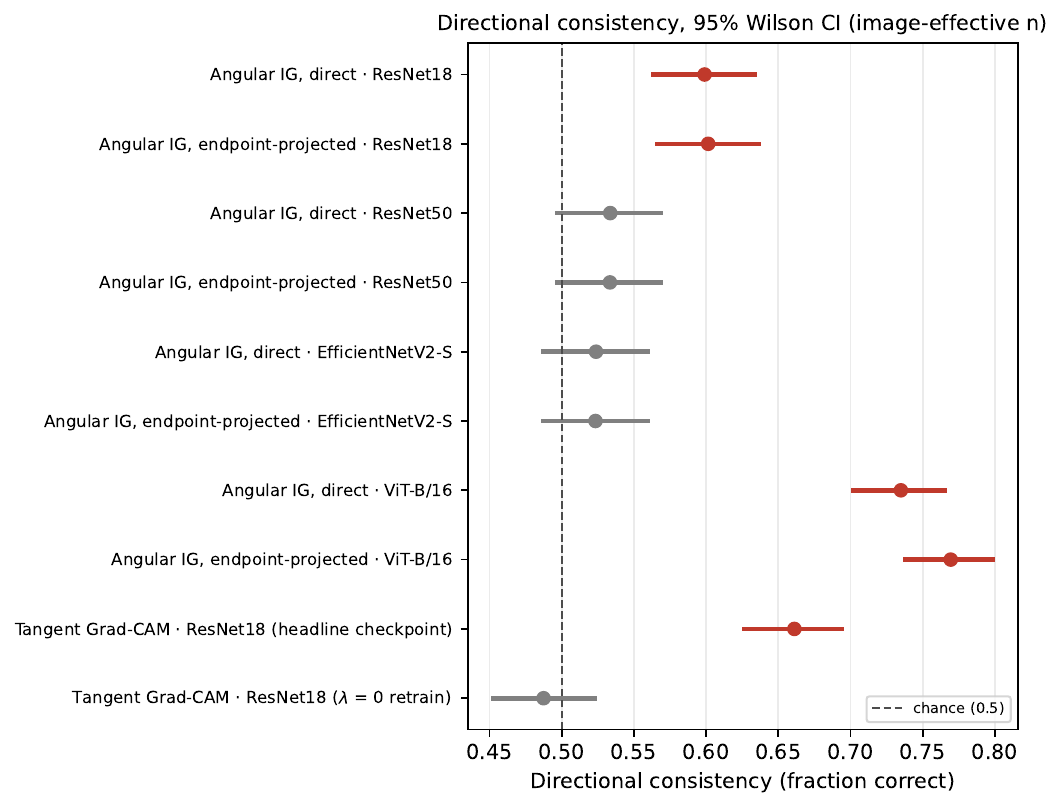}
\caption{Directional consistency with approximate image-effective Wilson intervals, plant $S^{1}$ (zero-fill deletion; blur IG integration baseline). Image-effective counts are 767 for Grad-CAM and 743 for angular IG; 24 images have no positive IG pixel and are skipped. The dashed line marks chance. Each image contributes multiple pixel deletions, so treating all deletions as independent understates uncertainty. Intervals use the number of contributing images as an approximate effective sample size; they are not calibrated clustered intervals and do not capture training variability. The two tangent Grad-CAM rows show different ResNet18 checkpoints, with DCS 66.1\% and 48.7\%. Estimates near 52--53\% should not be described as established above-chance performance under these intervals.}
\label{fig:8}
\end{figure*}

\begin{table*}[tp]
\centering
\caption{Top: angular versus standard IG deletion AUC (h), blur IG baseline, zero-fill deletion, 50 steps, $n=767$; $\Delta$ is direct AIG minus the best standard target selected on this split. Bottom: attribution-sum diagnostics on ResNet18, $n=50$. Ratios are medians over images with $|T|>10^{-3}$. Direct AIG uses wrapped angular change in radians, equal to lifted change here because no path winds; its mean absolute error is 0.0186 rad (0.071 h). Endpoint-projected AIG uses $T=rr'\sin\Delta\hat\theta$, in squared raw-output units. Standard rows sum absolute attributions and divide by their own scalar change, in raw-output units; they are scale diagnostics and test neither completeness nor its failure.}
\label{tab:6}
\small
\setlength{\tabcolsep}{4pt}
\begin{tabularx}{\linewidth}{@{}lXccc@{}}
\toprule
Backbone & AIG (tan) & AIG (dir) & Std IG (best) & $\Delta$ \\
\midrule
ResNet18 & 6.166 & 6.157 & 6.164 & $-0.007$ \\
ResNet50 & 7.078 & 7.065 & 7.064 & $+0.001$ \\
EfficientNet & 6.333 & 6.360 & 6.344 & $+0.016$ \\
ViT & 8.397 & 8.394 & 8.427 & $-0.033$ \\
\midrule
IG variant & Target $T$ / units & Mean abs err & Median ratio & \\
\midrule
Direct & wrapped $\Delta\theta$ (rad) & \textbf{0.0186} & \textbf{1.008} & \\
Endpoint-projected & $rr'\sin\Delta\hat\theta$ (raw-output units squared) & \textbf{0.012} & \textbf{1.018} & \\
Standard (scale only) & $\Delta(\hat y_{\sin}+\hat y_{\cos})$ (raw output) & 17.79 & $-45.17$ & \\
Standard (scale only) & $\Delta\|\hat y\|$ (raw output) & 6.16 & 23.42 & \\
\bottomrule
\end{tabularx}
\end{table*}

\begin{figure*}[tp]
\centering
\includegraphics[width=\linewidth]{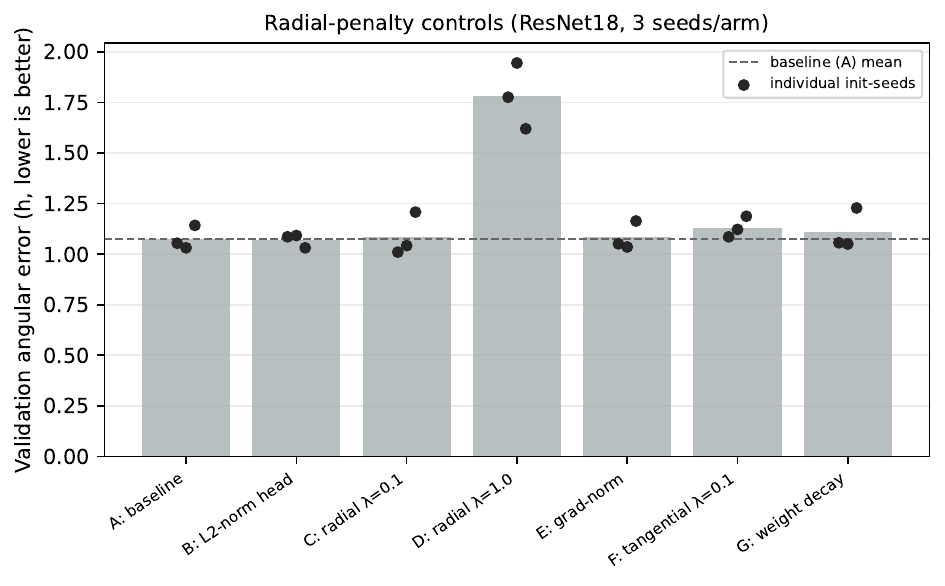}
\caption{Radial-penalty controls (ResNet18, plant $S^{1}$), normalised scale-invariant angular metric; bars are means over three initialization seeds with individual seeds overlaid. Every arm except the strong radial penalty (D) lies within seed-level variation of the baseline (A); $D$ has higher rescored error, with checkpoint selection confounded by the earlier norm-dependent metric (Appendix~\ref{sec:O.2}).}
\label{fig:9}
\end{figure*}

\textbf{Centred spherical occlusion.} On the same original checkpoint and the same 2,996 p13 images, patch 32, stride 16 and overlap-add over each patch footprint (169 forward passes) give $9.53/10.80/11.10/11.22/11.28/11.35/11.39/11.37/11.15^\circ$ at fractions 0.1--0.9. The mean is $11.02^\circ$, versus $9.37^\circ$ for the earlier estimator, and exceeds tangent Grad-CAM at every fraction. Both estimators use a black occluder and unsigned geodesic displacement; the other three arms in the validation comparison reproduced to the reported precision.

\textbf{Real $S^{1}$: additional method details.} Table~\ref{tab:7} gives the complete set for the four validation checkpoints. The unrectified tangent CAMs coincide on the CNNs with global average pooling (GAP); on ViT, tangent HiResCAM and Grad-CAM differ. The two-polarity tangent LayerCAM runs include standard ambient-sum LayerCAM as an in-run control, reproducing its scores to two decimals. These controls do not by themselves make polarity aggregation identical across methods. Standard and unsigned-geodesic occlusion score $6.85/7.63/6.86/8.11 h$ and $6.86/7.62/6.86/8.11 h$, respectively. The signed-geodesic scores are lower $(4.74/5.49/4.97/7.80 h)$, consistent with cancellation in overlap averaging. Occlusion is the strongest perturbation method in this evaluation; standard IG is higher on ViT. Directional results are shown in Figure~\ref{fig:8} and discussed below.

\begin{table*}[tp]
\centering
\caption{Plant $S^{1}$ deletion AUC (hours; mean geodesic displacement of the decoded prediction over the nine deletion fractions under the normalised metric of \S\ref{sec:4}, zero-fill deletion, blur IG baseline, single checkpoint per backbone, $n=767$ validation images also used for checkpoint selection; scalar-angle and circular LIME with the same von Mises weights ($\kappa=5$, 1000 perturbations), angular IG with 50 steps). Rows are grouped by CAM construction. The two-polarity tangent LayerCAM variant changes both the target and polarity aggregation relative to standard ambient-sum LayerCAM. The 'Invariant' column marks methods that satisfy the decoder-null invariance condition of Proposition~\ref{prop:1} ($\checkmark$\textsuperscript{*} = invariant to a global rescaling of $z$ only, \S\ref{sec:3.2}; tangent angular IG is not invariant, Proposition~\ref{prop:3}). Values in parentheses are directional consistency (DCS) for the signed methods: the fraction of the top-100 positively attributed single-pixel deletions that move the prediction against the attributed direction, as removal should (chance 0.5). Tangent Grad-CAM, HiResCAM and LayerCAM coincide on the three GAP CNNs (Appendix~\ref{sec:H}); on $S^{1}$ the tangent LayerCAM entry is the tangent HiResCAM map under another name (\S\ref{sec:3.1}). The rectified tangent row is unsigned and was scored in separate runs with ambient-sum LayerCAM as an in-run control (in-run controls $5.06/4.45/3.53/0.75h$, reproducing the ambient-sum LayerCAM row). Occlusion rows use patch 16, stride 8, black fill and centred overlap-add accumulation (729 forward passes per image). The signed geodesic row is the map that carries a direction and a DCS; the unsigned geodesic row ranks by the same displacement without sign cancellation across overlapping patches; unwrapped angular occlusion ranks by the unwrapped decoded-angle difference, which is decoder-invariant but branch-dependent. KernelSHAP and RISE were scored on $S^{1}$ only before the metric correction and are not reported here; their spherical forms are in Table~\ref{tab:5}.}
\label{tab:7}
\small
\setlength{\tabcolsep}{4pt}
\begin{tabularx}{\linewidth}{@{}Xccccc@{}}
\toprule
Method & Invariant & ResNet18 & ResNet50 & EfficientNetV2-$S$ & ViT-$B/16$ \\
\midrule
Ambient-sum Grad-CAM & $\times $ & 2.10 & 0.94 & 0.65 & 1.10 \\
Ambient-sum HiResCAM & $\times $ & 2.11 & 0.94 & 0.65 & 0.74 \\
Tangent Grad-CAM & $\checkmark $ & 1.58 (0.661) & 1.32 (0.493) & 1.94 (0.454) & 0.73 (0.519) \\
Tangent HiResCAM (= tangent LayerCAM) & $\checkmark $ & 1.58 (0.661) & 1.32 (0.493) & 1.94 (0.454) & 1.46 (0.535) \\
Ambient-sum LayerCAM (rectified) & $\times $ & 5.06 & 4.45 & 3.53 & 0.75 \\
Rectified tangent LayerCAM & $\checkmark $ & 5.24 & 4.53 & 4.33 & 1.44 \\
Scalar-angle LIME & $\times $ & 2.00 & 2.23 & 2.45 & 4.02 \\
Circular LIME & $\checkmark $\textsuperscript{*} & 2.32 (0.587) & 2.36 (0.493) & 2.68 (0.449) & 4.78 (0.490) \\
Angular occlusion (unwrapped, unsigned) & $\checkmark $ & 6.85 & 7.63 & 6.86 & 8.11 \\
Circular occlusion (geodesic, unsigned) & $\checkmark $ & 6.86 & 7.62 & 6.86 & 8.11 \\
Circular occlusion (geodesic, signed) & $\checkmark $ & 4.74 (0.593) & 5.49 (0.548) & 4.97 (0.536) & 7.80 (0.558) \\
Scalar-target IG (best reported target) & $\times $ & 6.16 & 7.06 & 6.34 & 8.43 \\
Angular IG, tangent & $\times $ & 6.17 (0.601) & 7.08 (0.533) & 6.33 (0.523) & 8.40 (0.769) \\
Angular IG, direct & $\checkmark $ & 6.16 (0.599) & 7.07 (0.534) & 6.36 (0.524) & 8.39 (0.735) \\
\bottomrule
\end{tabularx}
\end{table*}

\emph{Single-point tangent-CAM direction varies across checkpoints.} ResNet18 DCS is 66.1\% in the headline run and 48.7\% in the $\lambda =0$ retrain, while the other backbones give values near or below 0.5. The high value from one ResNet18 checkpoint is therefore not a stable architecture-level result.

\emph{Path integration and direction.} For direct AIG with blur, the four DCS estimates are 59.9\%, 53.4\%, 52.4\%, and 73.5\%. The endpoint-projected AIG row instead gives 60.1\%, 53.3\%, 52.3\%, and 76.9\%. Both sets are numerically above 0.5, but the smaller excesses are within the approximate image-effective intervals in Figure~\ref{fig:8}. Black-baseline comparisons cover four backbones; a distributional baseline is reported only for ResNet18. Appendix~\ref{sec:E} reports direct-AIG ResNet18 DCS of 59.4\% with black, 60.1\% with grey, 59.9\% with blur, and 65.3\% with the distributional baseline. These are checkpoint-specific directional results, separate from the completeness identity.

\textbf{Standard-IG target selection.} Exactly two scalar targets were evaluated: the coordinate sum and output norm. ``Best'' selects the larger deletion AUC per backbone on the same 767-image validation split, post hoc. Sum/norm scores are 6.149/6.164, 7.064/7.062, 6.342/6.344 and 8.427/8.424 h for ResNet18, ResNet50, EfficientNet and ViT. Thus norm is selected for ResNet18 and EfficientNet, sum for ResNet50 and ViT; the candidates differ by at most 0.015 h.

\textbf{Angular IG and circular surrogates.} Direct and standard scalar-target IG have similar unsigned deletion scores: the largest absolute difference in Table~\ref{tab:6} is 0.033 h. A ResNet18 distributional-baseline comparison is also close (4.36 versus 4.35 h). Their accounting targets differ. Direct AIG sums to lifted angular change, endpoint-projected AIG to $rr'  \sin  \Delta \hat{\theta }$, and scalar-target IG to its scalar's change. The standard rows of Table~\ref{tab:6} instead sum absolute attributions and compare them with their own scalar change. These unsigned diagnostics are reported for scale only and cannot establish completeness or its failure.

\emph{Circular surrogates.} Circular LIME has higher deletion scores than scalar-angle LIME on all four checkpoints, by $0.32/0.14/0.24/0.76 h$. The corrected-metric $n=50$ kernel/surrogate ablation in Appendix~\ref{sec:F} gives higher circular-surrogate scores for each kernel, with a mean circular-surrogate gain of 0.16 h and a von Mises-versus-Euclidean kernel effect of $-0.07$ h. The previous $n=30$ pre-correction values are superseded by this stored run. This supports a contribution from the surrogate in that ablation, without establishing a general equivalence of kernels. $S^{1}$ KernelSHAP results were obtained only before the metric correction and are omitted.

\clearpage

\section{Full Seed-Replicated Gaze Table}\label{sec:B}

The complete nine-fraction version of Table~\ref{tab:2}.

\begin{table*}[tp]
\centering
\caption{MPIIGaze $S^{2}$ deletion, tangent vs ambient-sum Grad-CAM, five independent initialisations (ResNet18, $n=2,996$ images per run, identical LOPO split, validation participant p13 (also used for epoch selection), degrees, higher means greater deletion displacement). The Tangent and Ambient-sum columns give the mean across the five runs $\pm $ the SD across runs (not within-run image scatter), and $\Delta $ is the paired difference. Two uncertainty estimates are reported because they answer different questions. Seed columns give mean $\pm$ SD and an uncorrected paired $t$-test over the five runs (df=4), capturing training variability. The bootstrap column gives a 95\% percentile interval from resampling images (20,000 resamples); each image's paired difference is first averaged over the five seeds, so an image enters once and the five appearances of one image are not treated as independent. The seed and bootstrap point estimates coincide by construction; their intervals measure different sources of variability. They differ a great deal: at fraction 0.1 the seed-level 95\% $t$-interval half-width ($t_{0.975,4}$ $\times  0.47/\sqrt{5} \approx  0.58{}^\circ $) is about six times the bootstrap half-width $(\approx 0.10{}^\circ )$. Between-run variation dominates image-sampling variation here, so the seed columns are the conservative and controlling estimate. Deletion fractions are nested subsets, so these tests are dependent and descriptive rather than confirmatory. Relative improvements quoted in the text are means of per-seed ratios, not ratios of means.}
\label{tab:8}
\small
\setlength{\tabcolsep}{4pt}
\begin{tabularx}{\linewidth}{@{}Xccccc@{}}
\toprule
 & Deletion (deg) &  & Seed-level $(n=5)$ &  & Bootstrap (per-image) \\
\midrule
Fraction & Tangent & Ambient-sum & $\Delta  \pm $ SD & $p$ & $\Delta $ [95\% CI] \\
\midrule
0.1 & $6.50\pm 0.29$ & $4.52\pm 0.36$ & $+1.97\pm 0.47$ & \textbf{0.001} & $+1.97 [+1.88, +2.07]$ \\
0.2 & $8.07\pm 0.28$ & $5.80\pm 0.39$ & $+2.28\pm 0.47$ & \textbf{$<0.001$} & $+2.28 [+2.16, +2.40]$ \\
0.3 & $8.77\pm 0.36$ & $6.73\pm 0.52$ & $+2.03\pm 0.49$ & \textbf{0.001} & $+2.03 [+1.90, +2.17]$ \\
0.4 & $9.15\pm 0.69$ & $7.43\pm 0.59$ & $+1.72\pm 0.58$ & \textbf{0.003} & $+1.72 [+1.59, +1.86]$ \\
0.5 & $9.49\pm 1.00$ & $8.06\pm 0.61$ & $+1.43\pm 0.75$ & \textbf{0.013} & $+1.43 [+1.31, +1.55]$ \\
0.6 & $9.73\pm 1.16$ & $8.86\pm 0.75$ & $+0.87\pm 0.84$ & 0.081 & $+0.87 [+0.77, +0.98]$ \\
0.7 & $9.88\pm 1.23$ & $9.81\pm 1.00$ & $+0.06\pm 0.83$ & 0.870 & $+0.06 [-0.01, +0.14]$ \\
0.8 & $10.00\pm 1.26$ & $10.47\pm 1.31$ & $-0.48\pm 0.94$ & 0.321 & $-0.48 [-0.53, -0.42]$ \\
0.9 & $10.16\pm 1.24$ & $10.51\pm 1.29$ & $-0.35\pm 0.88$ & 0.416 & $-0.35 [-0.39, -0.32]$ \\
Mean & $9.08\pm 0.72$ & $8.02\pm 0.65$ & $+1.06\pm 0.60$ & \textbf{0.017} & $n/a$ \\
\bottomrule
\end{tabularx}
\end{table*}

\textbf{Bootstrap versus seed-level uncertainty.} At three seeds, the seed-level and per-image bootstrap estimates disagreed at fraction 0.7: the bootstrap interval excluded zero while the seed-level test did not come close. At five seeds they agree: the bootstrap interval is $+0.065{}^\circ  [- 0.010, +0.139]$ and the seed-level test gives $p=0.870$ ($2/5$ seeds favouring tangent), and the advantage has ended by 0.7 on both readings. We note the earlier disagreement because it is the pattern to expect whenever a large image sample is applied to a difference of $0.09{}^\circ $, far below any angular difference that would matter in use, and because it resolved with more training runs rather than more images. At the 10\% fraction the seed-level 95\% $t$-interval half-width $(\approx 0.58{}^\circ )$ is about six times the bootstrap half-width $(\approx 0.10{}^\circ )$, confirming training variability rather than image sampling is the binding source of uncertainty. We flag this because two results in this paper were overstated by a related mechanism: the single-checkpoint value in \S\ref{sec:7.1}, and the $S^{1}$ directional result that did not survive re-initialisation (\S\ref{sec:7.2}).

\section{Regression Accuracy}\label{sec:C}

\begin{table*}[tp]
\centering
\caption{Plant $S^{1}$ regression accuracy on the held-out test split (angular error in hours, $n=377$, normalised metric). Best-epoch selection for these checkpoints used the earlier unnormalised validation metric (\S\ref{sec:8}), so validation-best values are not reported; all test values are scored with the normalised metric.}
\label{tab:9}
\small
\setlength{\tabcolsep}{4pt}
\begin{tabularx}{\linewidth}{@{}Xccccc@{}}
\toprule
Backbone & Test mean & Test med. & $\le 0.5h$ & $\le 1h$ & $\le 2h$ \\
\midrule
ResNet18 & 1.48 & 0.81 & 34.5\% & 60.7\% & 82.8\% \\
ResNet50 & 1.33 & 0.69 & 39.5\% & 62.9\% & \textbf{86.5\%} \\
EfficientNetV2-$S$ & \textbf{1.30} & \textbf{0.64} & \textbf{41.1\%} & \textbf{65.5\%} & 83.8\% \\
ViT-$B/16$ & 1.42 & 0.73 & 39.0\% & 60.7\% & 84.4\% \\
Random guess & 6.00 & $n/a$ & $n/a$ & $n/a$ & $n/a$ \\
Constant (mean time) & 2.32 & $n/a$ & $n/a$ & $n/a$ & $n/a$ \\
\bottomrule
\end{tabularx}
\end{table*}

Gaze $S^2$ (MPIIGaze, ResNet18, LOPO): the original unseeded checkpoint (\texttt{gaze-train-r18-lopo}), used for Table~\ref{tab:5} and Figure~\ref{fig:5} left, has p13 validation error $5.49^\circ$ and p14 test mean $5.20^\circ$ ($n=3,000$), median $4.90^\circ$, 50.9\% within $5^\circ$ and 94.5\% within $10^\circ$. The five seeded models of Table~\ref{tab:8} have p13 validation errors $5.63/5.29/5.36/5.38/5.23^\circ$ and no test evaluation. All reported gaze attribution uses p13 ($n=2,996$), also used for epoch selection.
\section{Entangled Oracle Control}\label{sec:D}

Table~\ref{tab:10} reports the entangled single-stream oracle control.

\begin{table*}[tp]
\centering
\caption{Entangled single-stream oracles ($n=200$ each). Mean AMR (std). Grad-CAM and HiResCAM variants are numerically close within each manifold; Saliency varies more across projections.}
\label{tab:10}
\small
\setlength{\tabcolsep}{4pt}
\begin{tabularx}{\linewidth}{@{}Xcc@{}}
\toprule
Method & $S^{1}$ & $S^{2}$ \\
\midrule
Model angular error & $0.364{}^\circ $ & $4.667{}^\circ $ \\
Ambient-gradient-magnitude Grad-CAM & 0.804 (0.081) & 0.988 (0.006) \\
Tangent Grad-CAM & 0.802 (0.079) & 0.988 (0.007) \\
Radial Grad-CAM & 0.807 (0.083) & 0.985 (0.014) \\
Ambient-gradient-magnitude HiResCAM & 0.663 (0.059) & 0.721 (0.020) \\
Tangent HiResCAM & 0.667 (0.060) & 0.722 (0.022) \\
Radial HiResCAM & 0.660 (0.060) & 0.720 (0.027) \\
Ambient-gradient-magnitude Saliency & 0.464 (0.082) & 0.611 (0.051) \\
Tangent Saliency & 0.504 (0.092) & 0.629 (0.057) \\
Radial Saliency & 0.437 (0.082) & 0.573 (0.073) \\
\bottomrule
\end{tabularx}
\end{table*}

\section{Baseline Ablations}\label{sec:E}

Under a black baseline on pre-masked images, $(x- x' ) \approx  x$, assigning zero attribution to about 90\% of pixels by construction and inflating deletion AUC for the IG family relative to gray and blur baselines (ResNet18 direct AIG deletion AUC under the normalised metric: 7.79h black, 0.91h gray, 6.16h blur, 4.36h distributional). The gray value collapses for the mirror-image reason: on a pre-masked image a gray baseline makes $(x- x' ) \approx  - 0.5$ over the whole blanked background, so IG spreads attribution across background pixels whose replacement by a black deletion baseline leaves their values unchanged. Directional consistency is far more stable across baselines (ResNet18 direct AIG: 59.4\% black, 60.1\% gray, 59.9\% blur, 65.3\% distributional), which is why we treat DCS as the primary evaluation metric for signed attribution and blur as the default IG integration baseline. Deletion sets selected pixels to zero for every row of this ablation.

The reference images, in raw $[0,1]$ space, are black ($x'=0$), gray ($x'=0.5$), Gaussian blur (kernel size 29, $\sigma=29$ pixels), and distributional (IG averaged over reference images drawn from other validation images). The 89--90\% black-pixel fraction was measured on 300 random masked images at $224\times224$.
\section{Circular Surrogates and Kernel Ablation}\label{sec:F}

\subsection{Surrogate construction}\label{sec:F.1}

We replace LIME's Euclidean surrogate with a circular-linear one, two ridge regressions $\hat{y}_{\sin }(b) = \beta _{\sin }^{\top }b$ and $\hat{y}_{\cos }(b) = \beta _{\cos }^{\top }b$ over superpixel indicators $b \in  \{0,1\}^{k}$. The decoded angle $\operatorname{atan2}(\hat{y}_{\sin }(b), \hat{y}_{\cos }(b))$ wraps correctly by construction, and the per-superpixel importance is $\phi _{j} = \hat{y}^{(0)}_{\cos } \beta _{\sin ,j} -  \hat{y}^{(0)}_{\sin } \beta _{\cos ,j}$. The Gaussian locality kernel becomes a von Mises kernel $w_{i} = \exp (\kappa  \cos (\theta _{i} -  \theta _{0}))$, the maximum-entropy distribution on $S^{1}$ for a fixed first circular moment. The $S^{2}$ analogue regresses three coordinates and projects with $P_{p}$. The $S^1$ surrogate comparison evaluates LIME; scalar-angle LIME uses the same von Mises weights as circular LIME. The surrogates are fitted to the raw outputs and projected at the unperturbed prediction, so the construction is invariant to a global rescaling of $z$ but not to input-dependent rescaling across the perturbation set.

\textbf{Spherical KernelSHAP.} The baseline in Table~\ref{tab:5} uses coarse quickshift superpixels (kernel size 8, maximum distance 400, ratio 0.2; approximately 40--70 segments). We sample $N=4,000$ Bernoulli$(1/2)$ coalitions, force the full and empty coalitions, and replace masked superpixels by the image mean colour. Three Shapley-weighted ridge regressions (penalty 1) fit the coordinates of the unit-vector prediction. At the unperturbed prediction $p_0$, the unsigned superpixel importance is $\alpha_j=\|P_{p_0}\beta_{:,j}\|_2$, with $P_{p_0}=I-p_0p_0^\top$. Unlike the raw-output circular surrogate, this baseline fits the normalised prediction.

\subsection{Kernel and surrogate ablation}\label{sec:F.2}

Table~\ref{tab:11} reports the LIME kernel and surrogate ablation.

\begin{table*}[tp]
\centering
\caption{Corrected-metric LIME ablation, ResNet18, $n=50$ validation images, $\kappa=5$, 1,000 perturbations with a shared per-image random stream, mean-colour superpixel masking, zero-fill deletion, normalised metric. Values are means and per-image SD. The main comparison in Table~\ref{tab:7} uses 767 images; absolute scores are not directly interchangeable.}
\label{tab:11}
\small
\setlength{\tabcolsep}{4pt}
\begin{tabularx}{\linewidth}{@{}XXcc@{}}
\toprule
Kernel & Surrogate & Deletion AUC (h) & std \\
\midrule
Euclidean & scalar & 2.501 & 1.978 \\
von Mises & scalar & 2.362 & 1.952 \\
Combined & scalar & 2.387 & 1.892 \\
Euclidean & circular & \textbf{2.573} & 1.764 \\
von Mises & circular & \textbf{2.575} & 1.847 \\
Combined & circular & \textbf{2.590} & 1.854 \\
\bottomrule
\end{tabularx}
\end{table*}

\section{Insertion}\label{sec:G}

\textbf{Insertion.} On ResNet18 ($n=50$, blur), insertion AUC spans 1.7950--1.8696h across all six CAM methods, a 0.07h range, with every tangent/standard pair separated by $<0.001h$. The metric does not discriminate between these methods on this task and we draw no conclusions from it.

\section{\texorpdfstring{Grad-CAM $\equiv $ HiResCAM Under GAP}{Grad-CAM = HiResCAM Under GAP}}\label{sec:H}

\begin{table}[tp]
\centering
\caption{Grad-CAM vs. HiResCAM divergence ($\Delta $, hours, normalised metric).}
\label{tab:12}
\small
\setlength{\tabcolsep}{4pt}
\begin{tabularx}{\linewidth}{@{}Xcc@{}}
\toprule
Backbone & Amb. $\Delta $ & Tangent $\Delta $ \\
\midrule
ResNet18 & 0.002 & 0.000 \\
ResNet50 & 0.000 & 0.000 \\
EfficientNetV2-$S$ & 0.000 & 0.000 \\
ViT-$B/16$ & 0.359 & 0.725 \\
\bottomrule
\end{tabularx}
\end{table}

The ambient-sum column reproduces Draelos \& Carin \cite{draelos2021}, whose \S3.6.2 proof covers GAP followed by a single fully connected layer. Our head is $\operatorname{Linear}(d,2)+\tanh $, which differs in two ways, and the equivalence survives both. The tanh is elementwise and applied after the linear layer, contributing a per-output scalar factor that is spatially constant and cancels under normalization. More substantively, our attribution is the tangent projection (the coefficient in \S\ref{sec:3.1}), a linear combination of two output gradients whose coefficients $(\hat{y}_{\cos }, -\hat{y}_{\sin })$ are per-image constants, so spatial constancy is preserved and the collapse extends to the tangent-projected variants, which their proof does not cover.

The new observation is the ViT column: CLS-token readout breaks spatial constancy, and the tangent projection \emph{widens} the divergence (0.725 vs. 0.359h), so the choice of CAM variant matters roughly twice as much for geometry-aware attribution on non-GAP models. Under these GAP-head assumptions, comparing the two maps does not distinguish attribution targets. LayerCAM is not covered by this collapse: its per-location rectification $\operatorname{ReLU}(g)$ is applied before the channel sum, and $\operatorname{ReLU}(\sum _{k} w_{k} A_{k}) \ne  \operatorname{ReLU}(\sum _{k} \operatorname{ReLU}(w_{k}) A_{k})$ in general, which is why standard LayerCAM differs from standard Grad-CAM by 2.88--3.51 h on the three CNNs (Table~\ref{tab:7}). The tangent LayerCAM entries coincide with tangent HiResCAM because no rectification is applied to the projected field (\S\ref{sec:3.1}).

\section{Additional Diagnostics and Training Controls}\label{sec:I}

\begin{table*}[tp]
\centering
\caption{Direction-specificity across backbones (validation error, hours, mean $\pm $ SD over 3 init-seeds, normalised scale-invariant metric; checkpoints were selected under the earlier unnormalised metric).}
\label{tab:13}
\small
\setlength{\tabcolsep}{4pt}
\begin{tabularx}{\linewidth}{@{}Xccc@{}}
\toprule
Backbone & Baseline (A) & Radial penalty (C) & Tangential penalty (F) \\
\midrule
ResNet18 & $1.076 \pm $.058 & $1.087 \pm $.107 & $1.131 \pm $.052 \\
ResNet50 & $1.009 \pm $.060 & $1.020 \pm $.027 & $1.042 \pm $.022 \\
ViT-$B/16$ & $0.944 \pm $.066 & $0.940 \pm $.023 & $0.889 \pm $.030 \\
\bottomrule
\end{tabularx}
\end{table*}

Under the normalised metric the three arms lie within seed-level variation of one another on every backbone, and on ViT the tangential-penalty arm is numerically the best: there is no direction-specific asymmetry between penalising the radial and the tangential fraction.

\textbf{Effect on attribution.} Training with the radial penalty changes tangent-CAM directional consistency non-monotonically: ResNet18 48.7\% / 74.0\% / 41.8\% at $\lambda  = 0 / 0.1 / 1.0$, ResNet50 $48.8 / 55.5 / 55.2$, ViT $54.1 / 50.0$ / degenerate, the last with a flat deletion curve that leaves DCS undefined. A moderate penalty can make sign informative on ResNet18, but the pattern does not hold at higher $\lambda $ or across backbones, and the ViT collapse shows a strong penalty can produce a model that is accurate yet has no usable attribution structure. We regard this as suggestive and under-powered at one seed per cell.

For the permutation test, the response is the tangent-minus-ambient-gradient-magnitude saliency AMR gap. We relabel this gap within each seed block, use $|r_{\mathrm{pooled}}|$ as a two-sided statistic, and compute $p=(k+1)/(N+1)$ for $N=200,000$ relabellings. All four predictors have $k=0$ (largest null $|r|=0.74$), giving $p=5.0\times10^{-6}$ to the reported precision. Here ``$\ker D\pi$ energy'' is the null-gradient fraction, ``map overlap'' is attribution-map cosine similarity, and ``probe $R^2$'' is angle-leakage $R^2$.

\begin{table*}[tp]
\centering
\caption{Repeated-measures statistics for representation-sharing predictors.  The table below reports $\alpha $ and the three model-measured predictors introduced in \S\ref{sec:6.3}. For each it gives the pooled correlation (as in the predictor analysis), the correlation across the 7 condition means, a seed-level slope test ($t$, df=4), and a permutation test preserving seed structure (200,000 within-seed relabelings). Condition-mean correlations are consistently stronger than pooled, since averaging over seeds removes between-seed intercept noise.}
\label{tab:14}
\small
\setlength{\tabcolsep}{4pt}
\begin{tabularx}{\linewidth}{@{}Xcccc@{}}
\toprule
Predictor & Pooled $r$ & Condition-mean $r$ & Seed $t(4)$, $p$ & Permutation $p$ \\
\midrule
$\alpha  \to $ gap & $-0.86$ & $-0.997$ & $-7.49$, 0.0017 & $5.0\times10^{-6}$ ($k=0$) \\
$\ker  D\pi $ energy $\to $ gap & $+0.83$ & $+0.991$ & $+5.01$, 0.0075 & $5.0\times10^{-6}$ ($k=0$) \\
map overlap $\to $ gap & $-0.76$ & $-0.989$ & $-5.28$, 0.0062 & $5.0\times10^{-6}$ ($k=0$) \\
probe $R^{2} \to $ gap & $-0.69$ & $-0.941$ & $-5.64$, 0.0049 & $5.0\times10^{-6}$ ($k=0$) \\
\bottomrule
\end{tabularx}
\end{table*}

Of the four predictors, $\ker D\pi$ energy is the noisiest: it is monotone in $\alpha$ in only $2/5$ seeds, versus $4$--$5/5$ for the others.

A further candidate, a linear discriminant separating the two representations, does not track the gap: held-out accuracy stays at 1.000 for six of seven $\alpha $ values, uncorrelated with the gap $(r=+0.18, p=0.29)$. Two representations can be perfectly linearly separable while sharing all the information that matters, so linear separability is the wrong notion of entanglement here.

\textbf{Radial gradient fraction across $S^{1}$ backbones.} Measured over 150 validation images, $\rho _{\mathrm{rad}}$ is $0.228\pm 0.149$, $0.230\pm 0.163$, $0.225\pm 0.155$ and $0.239\pm 0.168$ for ResNet18, ResNet50, EfficientNet and ViT respectively, consistent with the 0.200 of the separately trained control arm (\S\ref{sec:8}): the estimates are similar across these checkpoints, so this backbone comparison does not establish predictive value for $\rho _{\mathrm{rad}}$.

\textbf{An under-powered comparison on $S^{2}$.} Gaze control arms use an unnormalised Linear$(512,3)$ head, unlike the hard-normalised headline models. Arms at $\lambda=0/0.1/1.0$ give best-epoch validation errors of $2.51{}^\circ $, $4.95{}^\circ $ and $3.91{}^\circ $, which would suggest the penalty is harmful on $S^{2}$. We draw no such conclusion. These are distinct training runs from the headline checkpoint with validation-best error $5.49{}^\circ $. The arms share the headline LOPO split (verified from the stored indices), but their validation curves are volatile enough that best-epoch values alone are difficult to interpret: the $\lambda =0$ curve spans 2.51--$12.80{}^\circ $ with a mean epoch-to-epoch change of $3.32{}^\circ $, larger than the $2.44{}^\circ $ cross-arm gap, and its $2.51^\circ$ minimum is the epoch-1 validation value on p13. Epoch-averaged errors are $7.33/8.38/7.31{}^\circ $; these single-run summaries do not establish an effect of the penalty. We treat them as under-powered and make no claim either way about whether the penalty transfers to $S^{2}$.

\textbf{The masking-boundary confound.} Plant pixels occupy about 1.3\% of a masked image on average, and 25--39\% of attribution mass falls on the boundary band rather than on plant interiors, suggesting the model may exploit the segmentation edge (which correlates with camera angle and hence time) as a shortcut. Retraining on aggressively edge-softened images did not degrade accuracy (1.29h vs. 1.41h, single seed each), so the sharp edge is not necessary. But every faithfulness measure collapsed (DCS for tangent Grad-CAM $0.487 \to  0.176$, with deletion AUC falling as well). This is genuinely ambiguous: softening removes the mask artifact but also blurs real leaf margins, which plausibly inform leaf angle. A drop in deletion AUC is consistent both with "the artifact carried the attribution" and with "we destroyed real signal." Distinguishing them requires a mask perturbation leaving leaf structure intact, which we have not built.

\subsection{What predicts it on a trained model}\label{sec:6.3}\label{sec:I.1}

Because $\alpha $ is experimentally controlled, we also examine quantities measurable in a trained model. The fraction of input-gradient energy in $\ker  D\pi $ correlates positively with the attribution gap (pooled Pearson $r=+0.83$). The held-out linear probe's $R^{2}$ for recovering angle from the null-pathway representation correlates negatively (pooled $r=- 0.69$) and rises from $- 0.05$ to 0.80 across the sweep. Overlap between tangent and radial attribution maps is a third predictor (pooled $r=- 0.76$; Figure~\ref{fig:2}, right). Appendix~\ref{sec:I} reports condition-mean correlations, seed-level slope tests, and permutation tests alongside these pooled coefficients. Linear discriminability of the two representations does not track the gap. On the four real $S^{1}$ backbones, measured radial gradient fractions are similar; these results do not establish a predictor across real architectures.

For each trained run, map overlap uses the first 40 test images. On the $64\times64$ grid, let $t_i=|\hat y_{\cos}g_{\sin,i}-\hat y_{\sin}g_{\cos,i}|$ and $r_i=|\hat y_{\sin}g_{\sin,i}+\hat y_{\cos}g_{\cos,i}|$. Flatten the maps and compute $\langle t,r\rangle/(\|t\|_2\|r\|_2)\in[0,1]$, with no other normalisation; the run's overlap is the mean over these 40 images. It is zero by construction at $\alpha=0$. The pooled correlation $-0.76$ uses all 35 runs, whereas the condition-mean correlation $-0.989$ uses the seven per-$\alpha$ means.

\subsection{Blind points of fixed scalar targets}\label{sec:6.4}\label{sec:I.2}

For the quaternion coordinate sum, $g_s(q)=1-a(q)^2$, where $a(q)=\|P_q\mathbf1\|/2$ and $a=0$ at $q=\pm(1,1,1,1)/2$. The alignment analysis uses a separately trained wide-range oracle: Euler angles are sampled from $U(-2.8,2.8)$ rad, rather than $U(-1.2,1.2)$ for Table~\ref{tab:1}; there are 6,000 training images and 16 epochs. Half of the 2,000 test images are rejection-sampled to $a(q)<0.45$, with the same stratification in training. Mean geodesic error is $22.7^\circ$. Table~\ref{tab:1}'s $n=200$ rotation column instead uses the narrow-range oracle and the full ten-method suite.

AMR is binned using edges $0,0.2,0.4,0.6,0.8,0.95,1$ and plotted at bin midpoints. The lowest bin $[0,0.2)$ contains 52 images and has sum-scalar target AMR 0.495 (approximately 0.50); this is not an evaluation at the exact critical point. AMR rises to approximately 0.96 at high alignment, while tangent AMR stays above 0.9997 (Spearman $\rho=+0.83$ between sum AMR and alignment).

For PASCAL3D+, per-image alignment and displacements are first averaged over the five seeds, which share the same validation split. Spearman correlation over the 4,886 resulting image observations is $\rho=-0.284$; stacking all $5\times4,886$ image-seed pairs would instead give $-0.177$. Lower alignment is associated with larger tangent-over-sum deletion gaps in all five seeds, but also with harder poses, so this association does not isolate the cause of the deletion gain.

\begin{figure*}[tp]
\centering
\includegraphics[width=\linewidth]{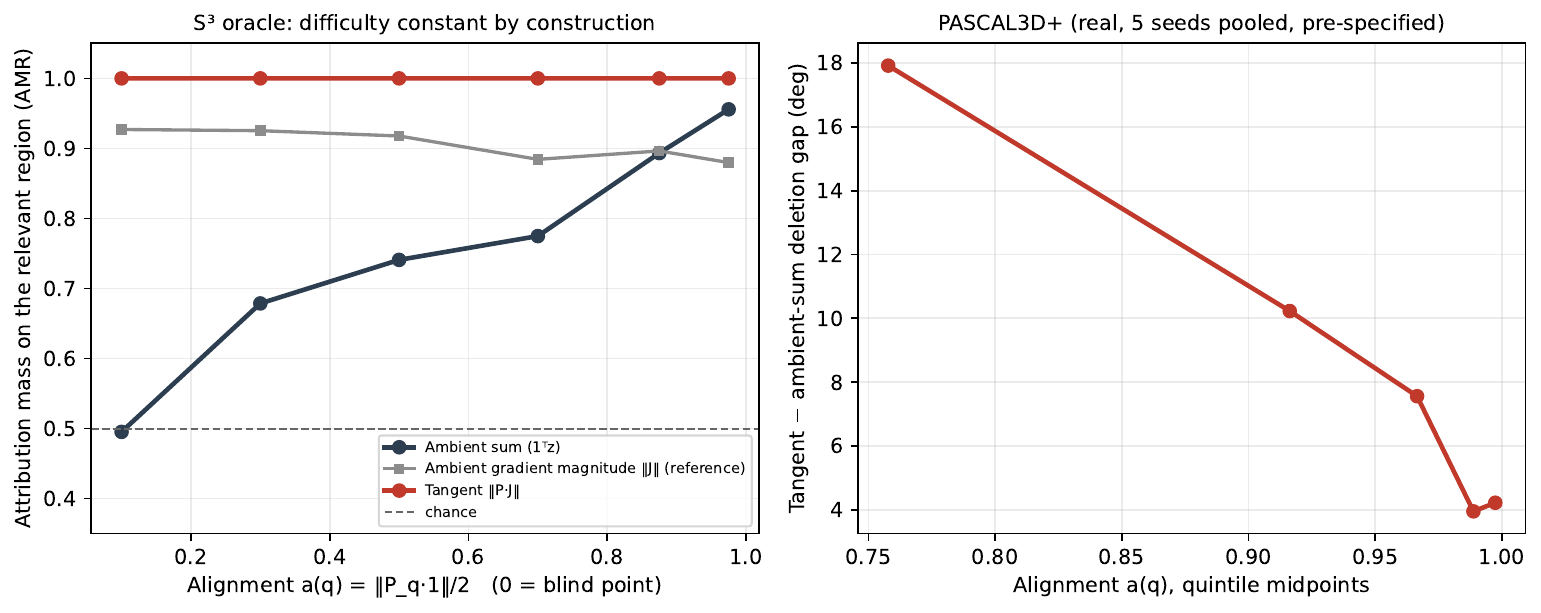}
\caption{Alignment and scalar attribution. Left: $S^{3}$ oracle AMR against alignment $a(q)=\| P_{q} 1\| /2$. Alignment zero identifies a critical point of the coordinate sum; bins have edges $0,0.2,0.4,0.6,0.8,0.95,1$ and are plotted at midpoints. The lowest bin contains 52 of 2,000 images and has AMR 0.495, while the tangent map remains near 1. Right: PASCAL3D+ validation data averaged per image over five seeds before binning, showing the tangent-over-sum deletion gap against alignment quintiles. The real-data association is confounded with pose difficulty; it does not measure the gap at an exact blind point.}
\label{fig:3}
\end{figure*}

\section{Plant Qualitative Comparison}\label{sec:J}

\begin{figure*}[tp]
\centering
\includegraphics[width=\linewidth]{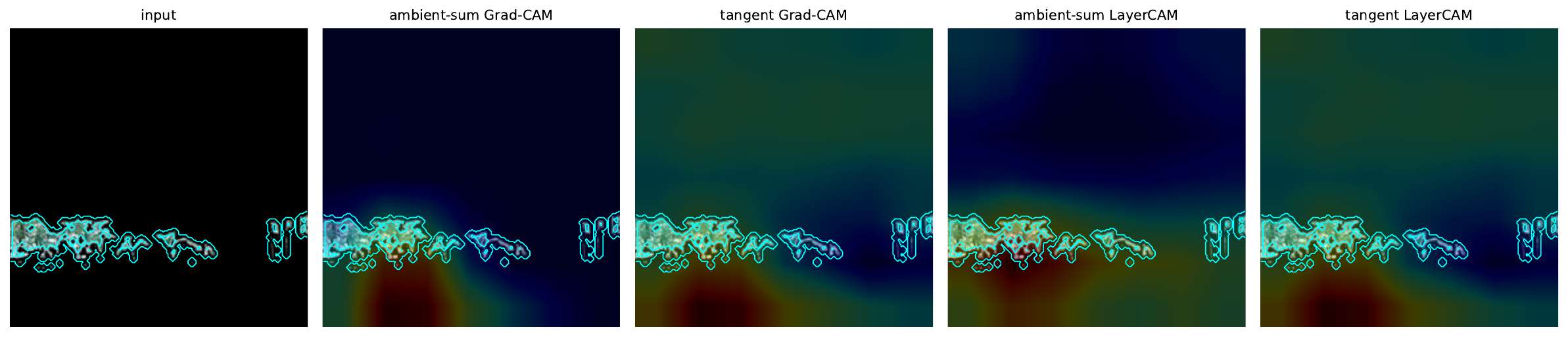}
\caption{Attribution on a representative masked plant image (ResNet18). Panels: input, ambient-sum Grad-CAM, tangent Grad-CAM, ambient-sum LayerCAM, tangent LayerCAM. The outlined band is the segmentation boundary region used in Appendix~\ref{sec:I}. Plant pixels occupy about 1.3\% of a masked image on average and 4.9\% of this image (\texttt{triton032\_6mm\_20250601\_215235\_420}) at $224\times224$, and no method concentrates attribution on plant interiors. Ambient-sum LayerCAM differs from ambient-sum Grad-CAM because LayerCAM rectifies each gradient before combining channels, which the GAP collapse of Appendix~\ref{sec:H} does not remove. The tangent LayerCAM panel is the tangent HiResCAM map under another name (no rectification is applied to the projected field, \S\ref{sec:3.1}), and under GAP that map coincides with tangent Grad-CAM, so the two tangent panels are identical by construction.}
\label{fig:10}
\end{figure*}

Figure~\ref{fig:10} shows the masked-plant attribution comparison.

\section{\texorpdfstring{Putting $S^{1}$ and $S^{2}$ Directional Metrics on a Common Ruler}{Putting S1 and S2 Directional Metrics on a Common Ruler}}\label{sec:K}

The $S^{1}$ and $S^{2}$ directional metrics in \S\ref{sec:4} have different scales. $S^{1}$ DCS is a binary sign-agreement rate with chance 0.5. $S^{2}$ uses a mean cosine in the two-dimensional tangent plane, with chance 0. The following conversion is an illustration under an assumed directional-error model, not a model-free comparison of task difficulty or method quality.

Assume an angular error variable $\phi $ follows a von Mises distribution with concentration $\kappa $. This is also the von Mises--Fisher model in ambient dimension $d_{\mathrm{vMF}}=2$. The probability of a correct half-plane sign and the mean directional cosine are

\begin{equation}
\begin{aligned}\Pr(\text{sign correct})&=\frac{\int_{-\pi/2}^{\pi/2}e^{\kappa\cos\varphi}\,d\varphi}{2\pi I_0(\kappa)},\\\mathbb E[\cos\varphi]&=\frac{I_1(\kappa)}{I_0(\kappa)}.\end{aligned}
\end{equation}

Both expressions increase with $\kappa$. The approximately 0.60 $S^1$ IG sign rate for ResNet18 implies $\kappa\approx0.319$. The $S^2$ tangent-CAM cosine 0.066 implies $\kappa\approx0.132$; comparing these gives a concentration ratio of about 2.4 but changes both method and task. The value 0.096 previously associated with IG is the tangent-CAM direction evaluated at IG-ranked pixels. The IG map itself is unsigned, so this proxy cannot support an IG-to-IG directional comparison.

For illustration, the $S^{2}$ tangent-CAM concentration would correspond to a sign rate of approximately 0.542, while the $S^{1}$ ResNet18 IG concentration would correspond to a cosine of approximately 0.157. This rescaling explains why the raw numbers 0.60 and 0.066 cannot be compared directly; it does not make their methods or tasks equivalent.

The assumed error model has not been validated here, and the $S^{1}$ reference is one ResNet18 checkpoint. The cross-task comparison is therefore descriptive. It does not isolate geometry from method, checkpoint, or task difficulty.

\section{Synthetic Oracle Construction}\label{sec:L}

All three oracle tasks share a common design: a $64\times 64$ single-channel image, values in [0,1], split vertically between columns 31 and 32 into a left half (columns 0--31, \emph{region A}) carrying the angular target and a right half (columns 32--63, \emph{region $B$}) carrying an irrelevant magnitude factor. Ground-truth attribution is region A by construction.

\subsection{\texorpdfstring{Region $B$ (all manifolds)}{Region B (all manifolds)}}\label{sec:L.1}

A fixed-phase sinusoidal grating of spatial frequency 3.0 cycles across the patch, amplitude-modulated by a contrast value $c \sim  \operatorname{Uniform}(0.3, 0.9)$, rendered as $\operatorname{clip}(0.5 + 0.5 c h, 0, 1)$. The contrast sets the \emph{norm} of the regression target and carries no angular information. Region $B$ is a grating rather than a noise field so that the two halves share spatial frequency, but the halves are not matched in image statistics: region A is phase-modulated and region $B$ contrast-modulated, and on $S^{2}$ region A is a sparse indicator. The guarantee that $B$ cannot change the decoded prediction is architectural (\S\ref{sec:L.6}) and does not depend on texture matching, so differences in AMR across manifolds should be read as descriptive rather than as evidence about texture. There is no additive noise anywhere in the generator.

\subsection{\texorpdfstring{Region A, $S^{1}$}{Region A, S1}}\label{sec:L.2}

A sinusoidal grating $h=\sin (3.0(0.7x + 0.3y) + \theta )$ mapped to [0,1] by $\tfrac12 (h+1)$, $i$.$e$. the phase offset \emph{is} the target angle, at a fixed diagonal orientation. Target angle $\theta  \sim  \operatorname{Uniform}(-\pi , \pi )$, regression target $c \cdot  (\sin \theta , \cos \theta )$, so the target's angle comes from region A and its norm from region $B$.

\subsection{\texorpdfstring{Region A, $S^{2}$}{Region A, S2}}\label{sec:L.3}

A bright indicator drawn radially from the image center at azimuth $\theta $, extending to 0.9 of the half-width in 80 steps with a small square stamp. Azimuth is encoded geometrically by the indicator's direction, elevation is encoded photometrically by its brightness, $0.4 + 0.6 \widetilde{e}$ where $\widetilde{e}$ is elevation rescaled to [0,1]. Sampling: azimuth $\sim  \operatorname{Uniform}(-\pi ,\pi )$, elevation $\sim  \operatorname{Uniform}(-\pi /3, \pi /3)$, bounded away from the poles for numerical stability. The unit vector is (cos e sin a, cos e cos a, sin e) and the target is $c \cdot  u$.

\subsection{\texorpdfstring{Region A, rotations $(S^{3})$}{Region A, rotations (S3)}}\label{sec:L.4}

Three horizontal bands of 21 rows each (the third absorbing the remainder), each a grating of frequency 3.0 at a distinct orientation ((0.7,0.3), (0.3,0.7), (0.5,0.5)) whose phase encodes one Euler angle. Convention is ZYX, converted to a (w,x,y,z) quaternion, each angle $\sim  \operatorname{Uniform}(-1.2, 1.2) \mathrm{rad}$. Target is $c$ times the unit quaternion, so the ambient output lies in $\mathbb{R} ^{4}$ and normalisation lands on $S^{3}$. The predicted quantity is the \emph{rotation} that quaternion denotes, and rotation error is measured with the $\operatorname{SO}(3)$ geodesic $2\arccos |\langle q,q'\rangle |$ accordingly. The higher ambient-gradient-magnitude AMR on this task is not explained by occupied area, since the $S^{1}$ grating also fills its half; we have not isolated its cause and treat it as descriptive.

\subsection{Rendering caveat}\label{sec:L.5}

Both the gratings and the $S^{2}$ indicator are rendered without anti-aliasing (direct evaluation on a pixel grid, integer rounding for the indicator). We do not expect this to affect the comparison, since all methods see identical images, but it is a difference from natural-image statistics worth stating.

\subsection{Two-stream architecture}\label{sec:L.6}

Each branch is $\operatorname{Conv}(1\to 16, 3\times 3) \to  \operatorname{ReLU} \to  \operatorname{MaxPool}(2) \to  \operatorname{Conv}(16\to 32, 3\times 3) \to  \operatorname{ReLU} \to  \operatorname{MaxPool}(2)$, giving a $32\times 16\times 8 = 4096$-dimensional flattened representation per branch. The angle branch applies a linear map to the ambient output dimension (2 for $S^{1}$, 3 for $S^{2}$, 4 for $S^{3}$) followed by $L^{2}$ normalization to a unit vector. The magnitude branch applies a linear map to a scalar followed by softplus. The prediction is the product, $\hat{y} = \operatorname{mag} \cdot  \hat{u}$.

\textbf{The branches receive strictly disjoint pixels}: the angle branch is given only columns 0--31 and the magnitude branch only columns 32--63, with no shared parameters or overlap. Consequently $\partial\hat u/\partial(\text{region B})$ $= 0$ identically, which is why exact ground truth is available and why a correctly-projected attribution should score 1.000 rather than merely high (\S\ref{sec:5}). Note that the oracle deliberately avoids GAP, so the GAP collapse of Appendix~\ref{sec:H} does not apply here.

\textbf{Ambient baseline as implemented on $S^{1}$. The ambient-gradient-magnitude Grad-CAM, HiResCAM and Saliency arms weight each activation (or input pixel) by $\sqrt{g_{\sin }^{2} + g_{\cos }^{2}}$, the magnitude of its two coordinate gradients, combined after two backward passes; the tangent arms use $\hat{y}_{\cos } g_{\sin } -  \hat{y}_{\sin } g_{\cos }$ and the radial arms $\hat{y}_{\sin } g_{\sin } + \hat{y}_{\cos } g_{\cos }$. A fourth arm back-propagating the true output norm $s(z) = \| z\| $, whose per-activation weight is $(\hat{y}_{\sin } g_{\sin } + \hat{y}_{\cos } g_{\cos })/\| z\| $, is the radial arm multiplied by a positive per-image factor and reproduces its attribution mass ratio exactly in a fresh run of the generator (norm vs radial AMR 0.092 vs 0.092 for Grad-CAM, 0.000 vs 0.000 for HiResCAM, 0.000 vs 0.000 for Saliency; this run is a separately trained instance, so its absolute values are not those of Table~\ref{tab:1}). The distinct ambient values in Table~\ref{tab:1} therefore come from the magnitude combination and not from a norm target.}

\textbf{Higher-dimensional oracle weights.} For $z\in\mathbb R^n$, $n=3,4$, let $p=z/\|z\|$, $P_p=I-pp^\top$, and $J_i=\partial z/\partial x_i$, obtained from $n$ backward passes. Ambient-gradient-magnitude, tangent and radial saliency use
\begin{equation}
\|J_i\|_2,\qquad \|P_pJ_i\|_2,\qquad |p^\top J_i|,
\end{equation}
respectively, with $\|J_i\|_2^2=\|P_pJ_i\|_2^2+(p^\top J_i)^2$. For quaternions, $P_q=P_{-q}$. CAM arms apply these formulas per activation before pooling. On $S^2/S^3$ these weights are nonnegative and the Grad-CAM ReLU is inert; the $S^1$ tangent and radial weights are signed.

The separate $S^1$ radial Grad-CAM AMR of 0.092 is a ratio artefact, not spatial leakage. On 46 of 500 images its ReLU zeros the entire region-$B$ map, leaving floating-point residue on region A. The per-image AMR is then 1, and approximately 0 on the other images (median $2.7\times10^{-7}$, SD 0.289, consistent with $\sqrt{0.092\times0.908}$). Branches are upsampled separately, so interpolation cannot move mass across the region boundary.

\subsection{Training}\label{sec:L.7}

MSE on the unnormalized output against $c \cdot  u$, so angle and magnitude are learned jointly. Adam, learning rate $10^{-3}$ fixed with no schedule, batch size 128, no weight decay, no augmentation. Training sets are 5,000 images per manifold, epochs are $8 (S^{1})$, $12 (S^{2})$, $14 (S^{3})$. Data RNG uses a fixed seed for train and seed+999 for test, so the two are drawn from the same generator with disjoint streams, model initialization and shuffling use a separate fixed seed. There is no validation split (these are oracles used to probe attribution, not models being selected) so all reported numbers are on the held-out test draw.

\subsection{Evaluation sets}\label{sec:L.8}

$n=500 (S^{1})$, $n=30 (S^{2})$, $n=200 (\operatorname{SO}(3))$. The $S^{2}$ figure is a deliberate reduction from the generator's default of 500 to bound the cost of the perturbation-based methods. It is the smallest evaluation in the paper and the corresponding SD should be read with that in mind.

\subsection{Entangled variant}\label{sec:L.9}

The control of \S\ref{sec:6} uses the same data generators but a single-stream model over the whole image: $\operatorname{Conv}(1\to 32) \to  \operatorname{ReLU} \to  \operatorname{Pool} \to  \operatorname{Conv}(32\to 64) \to  \operatorname{ReLU} \to  \operatorname{Pool} \to  \operatorname{Conv}(64\to 128) \to  \operatorname{ReLU} \to  \operatorname{AdaptiveAvgPool}(4) \to  \operatorname{Linear}$. The output head is a single linear map with no unit normalization and no separate magnitude branch, so angle and magnitude are fully entangled. Two differences from the oracle should be noted, since they are not purely a matter of merging the streams: the entangled model is deeper, and it uses adaptive average pooling where the oracle uses none. The control shows that the tangent CAM advantage is small in this entangled architecture, but does not isolate entanglement from depth and pooling as the cause. Grad-CAM is computed at the third convolutional layer. $\operatorname{SO}(3)$ is not covered by this control.

\subsection{Oracle AMR definition}\label{sec:L.10}

For an attribution map $\phi $,

\begin{equation}
\operatorname{AMR}=\frac{\sum_{(i,j)\in A}|\varphi_{ij}|}{\sum_{(i,j)\in A}|\varphi_{ij}|+\sum_{(i,j)\in B}|\varphi_{ij}|+\epsilon},
\end{equation}

with $\epsilon  = 10^{-12}$. Three details matter for interpretation. Mass is \emph{absolute value}, not a positive part: a method is credited for attributing to region A regardless of sign, which is the appropriate choice here because the signed methods are being evaluated for localization rather than direction. The region mask is purely geometric (column index) so it involves no thresholding or estimation. And normalization is per-image and applied \emph{after} masking, with the resulting ratios averaged over the test set, so no image dominates through map scale.

\subsection{Distinct measure of the same name}\label{sec:L.11}

The plant boundary analysis of Appendix~\ref{sec:I} uses a differently-defined quantity that we also called an attribution mass ratio. It shares the absolute-value convention but differs in every other respect: regions are derived by brightness-thresholding the masked image at 0.05 and applying morphological dilation and erosion with a 6-pixel radius to yield plant core, boundary band, and background, fractions are taken over the whole image and sum to one. It has no ground truth. Readers should not compare the two numerically, only the oracle version of the measure is validated against a known answer.

\section{Generalization to Higher-Dimensional Manifolds}\label{sec:M}

The tangent projection applies as stated to normalisation decoders $z \mapsto  z/\| z\| $ onto a sphere $S^{m} \subset  \mathbb{R} ^{m+1}$, where $\ker  D\pi $ is the single radial direction and coincides with the normal space of the sphere. For an embedded Riemannian $M \subset  \mathbb{R} ^{n}$ reached through a general decoder, $\ker  D\pi (z)$ is a subspace of the representation space that need not coincide with the normal space of $M$ (\S\ref{sec:2.1}), and attribution should pass through the full differential $D\pi $, as in the 6D arm of \S\ref{sec:7.3}. On $S^{2} (n=3, m=2)$ attribution becomes vector-valued with azimuthal and polar components. Our rotation oracle predicts a \emph{unit quaternion}, so its manifold is $S^{3} \subset  \mathbb{R} ^{4} (n=4, m=3)$: a single normal direction, the radial one, exactly as on $S^{1}$ and $S^{2}$. We report that representation's numbers throughout. The same group embedded as rotation matrices, $\operatorname{SO}(3) \subset  \mathbb{R} ^{3\times 3} \cong  \mathbb{R} ^{9} (n=9, m=3)$, would instead have six normal directions of nine. We did not run that experiment, and we state the contrast only to make the point that normal-space dimension is a fact about the parameterisation a practitioner chooses, not about the quantity being predicted.

\subsection{The double cover}\label{sec:M.1}

Unit quaternions cover $\operatorname{SO}(3)$ twice, with $q$ and -q denoting the same rotation. This identification is discrete, so it leaves $D\pi $ and therefore the tangent projection unchanged, but it does affect distance: every rotation error we report uses the double-cover-aware geodesic $2\arccos |\langle q,q'\rangle |$, which is invariant to the sign flip. An attribution method that ignored the identification would still compute the correct tangent directions, and a \emph{metric} that ignored it would report $2 \arccos (- 1) = 360{}^\circ $ for two identical rotations under the stated formula ($180{}^\circ $ as the $S^{3}$ geodesic without the factor of two), the rotation analogue of the wrap-around failure of \S\ref{sec:2}.

The oracle evaluations cover $S^{1}$, $S^{2}$, and unit-quaternion rotations; real-data evaluations cover circular regression, gaze, and 3-DoF rotation. Full 6-DoF pose is outside this study. The ratio $(n- m)/m$ describes a parameterisation rather than predicting empirical benefit. Radial sensitivity can explain oracle contamination, but the hard-normalised gaze and quaternion comparisons instead concern scalarisation loss. The controlled representation-sharing result has not been established as a predictor across real architectures.

\section{Deferred Proofs, Statements and Disclosures}\label{sec:N}

\subsection{Representation independence}\label{sec:N.1}

Let two models factor the same prediction map, $\pi _{1} \circ  f_{1} = \pi _{2} \circ  f_{2} = F$, with $f_{i}$: $\mathbb{R} ^{d} \to $ $\mathbb R^{n_i}$, where $n_{1}$ and $n_{2}$ may differ. Then, for every $x$, $D\pi _{1}(f_{1}(x))Df_{1}(x) = D\pi _{2}(f_{2}(x))Df_{2}(x) = DF(x)$, by the chain rule applied to each factorisation. The complete decoded input differential depends only on the function $x \mapsto  y$ the model computes, while the ambient derivatives $Df_{1}$ and $Df_{2}$ depend on how that prediction is represented and may live in different spaces. The statement concerns this local differential only. A scalar or matrix built from the output can annihilate the same kernel while rescaling or distorting the retained directions, and CAM maps computed at internal layers of two different networks need not coincide even when their input Jacobians do.

\subsection{Full statements and proofs for decoder invariance}\label{sec:N.2}

Write a structured-output model as a composition

\begin{equation}
x\longrightarrow z=f(x)\in\mathbb R^n\longrightarrow y=\pi(z)\in M,
\end{equation}

with $f$ the network, $z$ its ambient representation, $\pi $ the fixed decoder and $M$ a smooth Riemannian manifold ($S^{1}$, $S^{2}$, and $\operatorname{SO}(3)$ here, with $S^{3} \subset  \mathbb{R} ^{4}$ used as the unit-quaternion representation of rotations). We assume $\pi $ is smooth and a submersion on an open set $U \subseteq  \mathbb{R} ^{n}$ containing the image of $f$, so that $\ker  D\pi (z)$ has constant dimension $n -  \dim  M$ on $U$; atan2, $z \mapsto  z/\| z\| $ and the Gram-Schmidt decoder of \S\ref{sec:7.3} qualify on their natural domains. We do not assume $\| z\|  = 1$. In every case here $n$ exceeds dim M, so $\ker  D\pi (z)$ is nontrivial.

\textbf{Definition~\ref{def:1}} (Decoder-null direction). \emph{A direction $\delta  \in  \mathbb{R} ^{n}$ at $z$ is decoder-null if $\delta  \in  \ker  D\pi (z)$.}

\textbf{Decoder-null} sensitivity does not move the prediction. \emph{Let} $\pi $ be $C^{2}$ on $U$ and $\delta  \in  \ker  D\pi (z)$. Then $\pi (z + \epsilon \delta ) = \pi (z) + R(\epsilon )$ with $\| R(\epsilon )\|  \le  \tfrac12  \epsilon ^{2} \| \delta \| ^{2} \sup  \| D^{2}\pi \| $ over the segment; \emph{for the normalisation decoder $\| D^{2}\pi (z)\|  \le  3/\| z\| ^{2}$, and along the fibre itself the invariance is exact, $\pi (z + \epsilon z) = \pi (z)$ for $\epsilon  > - 1$. The first-order qualifier concerns input perturbations whose image is only tangent to the fibre, which is what the finite-deletion metrics of \S\ref{sec:7} probe.}

\textbf{Normalisation decoders.} \emph{For $\pi (z) = z/\| z\| $ and $z \ne  0$, with $p = z/\| z\| $,}

\begin{equation}
D\pi(z)=\frac1{\|z\|}(I-pp^\top)=\frac1{\|z\|}P_{T_p},
\end{equation}

\begin{equation}
\ker D\pi(z)=\operatorname{span}\{p\}.
\end{equation}

\emph{Proof.} $\partial _{j}(z_{i}/\| z\| ) = \delta _{ij}/\| z\|  -  z_{i}z_{j}/\| z\| ^{3} = (I -  pp^{\top })_{ij}/\| z\| $, and $I -  pp^{\top }$ annihilates $p$ and acts as the identity on $p^{\perp}$. \hfill$\square$

Changing only the magnitude of $z$ does not change the decoded prediction: the decoder keeps the direction and discards the scale (the representation independence of the full differential is stated in Appendix~\ref{sec:N}).

\textbf{Proposition~\ref{prop:1}} (Gauge dependence of ambient attribution). \emph{Let $\pi $ be the normalisation or the circle-valued angle decoder (represented locally by atan2), fix an input $x_{0}$ with $z = f(x_{0}) \ne  0$, and let $c$: $\mathbb{R} ^{d} \to  (0, \infty )$ be any smooth function with $c(x_{0}) = 1$. The network $\widetilde{f} = c\cdot f$ computes the same prediction map, $\pi  \circ  \widetilde{f} = \pi  \circ  f$, and for any differentiable scalar target $s$ the input gradient of the scalarised output at $x_{0}$ is}

\begin{equation}
\begin{aligned}\nabla_x(s\circ\widetilde f)(x_0)={}&\nabla_x(s\circ f)(x_0)\\&+\langle z,\nabla s(z)\rangle\nabla c(x_0).\end{aligned}
\end{equation}

\emph{The attribution of $s \circ  f$ at $x_{0}$ is independent of the representation, that is of $c$, if and only if $\langle z, \nabla s(z)\rangle  = 0$, equivalently $\nabla s(z) \in  (\ker  D\pi (z))^{\perp}$, which is also the condition for $s$ to be insensitive to decoder-null perturbations to first order. Every $s = h \circ  \pi $ satisfies it, since $\nabla (h \circ  \pi ) = D\pi ^{\top }\nabla h \in  (\ker  D\pi )^{\perp}$; conversely, if $\nabla s$ is horizontal on an open union of fibres then $s = h \circ  \pi $ there for a smooth $h$.}

Proof of Proposition~\ref{prop:1}. $J_{\widetilde{f}}(x_{0}) = c(x_{0}) J_{f}(x_{0}) + f(x_{0}) \nabla c(x_{0})^{\top } = J_{f}(x_{0}) + z \nabla c(x_{0})^{\top }$, so $\nabla _{x}(s \circ  \widetilde{f})(x_{0}) = J_{\widetilde{f}}(x_{0})^{\top } \nabla s(z) = J_{f}(x_{0})^{\top } \nabla s(z) + \nabla c(x_{0}) \langle z, \nabla s(z)\rangle $. Since $\nabla c(x_{0})$ can be any vector, the second term vanishes for every $c$ if and only if $\langle z, \nabla s(z)\rangle  = 0$, and $\ker  D\pi (z) = \operatorname{span}\{z\}$ by the normalisation Jacobian (\S\ref{sec:2.1}). For the last claim, $d/dt s(tz) = \langle z, \nabla s(tz)\rangle  = 0$ along each ray, so $s$ is constant on fibres; $\pi $ restricted to the set is a surjective submersion onto an open subset of $M$, so $h(p)$ :$= s(z)$ for any $z$ in the fibre over $p$ is well defined and, since $\pi $ admits smooth local sections, smooth. \hfill$\square$

The added term depends on the rescaling $c$ rather than on a change in the decoded prediction. We call it the gauge component of the rescaling response. For a nonzero scalar target gradient, the gauge fraction is

\begin{equation}
\begin{gathered}g_s(z)=\frac{\langle\nabla s(z),z\rangle^2}{\|\nabla s(z)\|^2\|z\|^2}\in[0,1],\\\nabla s(z)\ne0.\end{gathered}
\end{equation}

the fraction of squared \textbf{output-target gradient} norm in the radial direction. It is zero for decoder-invariant scalars wherever their gradient is nonzero; it is undefined when the gradient vanishes. This output-space fraction need not equal a fraction of input-gradient energy after multiplication by $J_{f}^{\top }$. Proposition~\ref{prop:2} concerns points of zero tangential sensitivity, including points with zero full gradient.

Tangent projection removes exactly these components. On $S^{1}$ the tangent coefficient used below equals $\| z\| ^{2}$ times the gradient of the decoded angle, a positive per-image factor, and the construction applies on manifolds with no single angle coordinate.

\textbf{Proposition~\ref{prop:2}} (Blind points of scalar targets). \emph{Let $\pi $ be the normalisation decoder onto $S^{n-1}$, $n\ge 2$, and let $s$: $\mathbb{R} ^{n}\setminus \{0\} \to  \mathbb{R} $ be smooth. For every $r>0$ there are at least two points $z$ with $\| z\| =r$ at which the tangential component of $\nabla s(z)$ vanishes. If $\nabla s(z)\ne 0$ at such a point, then $g_{s}(z)=1$ and the scalar gradient contains only radial sensitivity; if $\nabla s(z)=0$, the scalar input gradient vanishes. Nonzero linear scalar targets $s(z)=v^{\top }z$ attain the lower bound of two critical points, at $z=\pm rv/\| v\| $.}

\textbf{Proof of Proposition~\ref{prop:2}.} Restrict $s$ to $\{\| z\| =r\}$. At its maximum and minimum, the tangential derivative vanishes and $\nabla s$ lies in $\operatorname{span}\{z\}$. A nonconstant restriction has distinct extrema; a constant restriction has every point critical. If $\nabla s\ne 0$, only radial sensitivity remains; if $\nabla s=0$, the input gradient is zero. For $v\ne 0$, $v^{\top }z$ has exactly two critical points, $\pm rv/\| v\| $. \hfill$\square$

For the coordinate-sum scalar target on the quaternion sphere, $g_{s}(q)=1- a(q)^{2}$, where $a(q)=\| P_{q} 1\| /2$ is the alignment statistic used in \S\ref{sec:6.4}. Thus low alignment means that the fixed target direction retains little tangential sensitivity. Whether this produces a large AMR or deletion gap also depends on the model and heatmap construction. All nonzero linear scalar targets attain the same two-critical-point bound on the sphere; this does not rank their attribution quality. The full differential avoids loss of sensitivity from choosing one fixed scalar direction, although subsequent CAM aggregation can still cancel contributions (Appendix~\ref{sec:N}).

The normalisation Jacobian above restricts to a conformal isomorphism from the horizontal space to $T_{p} M$, with scale factor $1/\| z\| $. For the local $S^{1}$ CAM coefficient used below, output norm changes the map by a positive per-image factor; sign and rankings are therefore unchanged by that factor. This argument does not extend to path integration or surrogate fitting across inputs with different output norms. Metrics defined directly from unnormalised $z$ can also depend on the representation rather than only on $\pi (z)$, as Appendix~\ref{sec:O} shows.

\textbf{Consequences on the quaternion sphere and its double cover.} For the nonzero coordinate sum, the two critical points have measure zero. Its gauge fraction is $g_{s}(q)=1- a(q)^{2}$. On PASCAL3D+ validation crops, the minimum alignment per seed lies between 0.16 and 0.51, corresponding to 74--97\% of the fixed \textbf{output-target gradient} norm squared being radial at those extrema of the observed range. This is not a radial input-gradient fraction: the normalised head annihilates the radial component. Under uniform rotations, $g_{s}>\tfrac12 $ corresponds to two caps of angular radius $45{}^\circ $ with probability $2(\pi /4- \tfrac12 )/\pi \approx 18$\%. The observed poses are not uniform: median alignment is approximately 0.975, giving median $g_{s}\approx 5$\%. The observed alignment/deletion association remains confounded with pose difficulty.

Every nonzero linear scalar target has two critical points on the sphere; the count does not imply optimal deletion performance. A scalar that depends only on the rotation must satisfy $s(q)=s(- q)$ and descends to $\mathbb{R}  P^{3}\cong \operatorname{SO}(3)$. The Lusternik--Schnirelmann lower bound is four critical points on $\mathbb{R}  P^{3}$. In contrast, the full differential retains all local tangent directions without choosing one fixed scalar. Targets formed with a tangent frame are interpreted with that frame fixed at the current prediction; this statement concerns the differential and does not rule out cancellation in a derived CAM map.

\subsection{Direct angular IG: statement and proof}\label{sec:N.3}

Direct AIG applies standard IG to the decoded angle, integrating $\partial \hat{\theta }/\partial x_{i}$ along the straight path $\gamma (\tau ) = x'  + \tau (x -  x' )$ with $\hat{\theta }$ continuously lifted along it. Tangent AIG applies standard IG to each raw coordinate and combines the attributions at the endpoint, $\phi _{i}^{\mathrm{tan}} = \hat{y}_{\cos } \mathrm{IG}_{\sin ,i} -  \hat{y}_{\sin } \mathrm{IG}_{\cos ,i}$. The two produce correlated spatial maps (Pearson $r \approx  0.64$--0.84) but complete to different quantities and differ in their invariance (Proposition~\ref{prop:3}). The input baseline determines the integration path and its decoded trajectory: on pre-masked images a black baseline zeroes $(x -  x' )$ over about 90\% of pixels and inflates deletion scores. \textbf{Blur is our default $S^1$ integration baseline} (Appendix~\ref{sec:E}).

\textbf{Proposition~\ref{prop:3}} (Direct angular IG). \emph{Write $z(\tau ) = f(\gamma (\tau ))$ and assume $z(\tau ) \ne  0$ along the path. Let $\omega  = (z_{\cos } dz_{\sin } -  z_{\sin } dz_{\cos })/\| z\| ^{2}$ be the angle form on $\mathbb{R} ^{2}\setminus \{0\}$, so that $d\hat{\theta } = \omega $, and let $t(z) = (z_{\cos }, - z_{\sin })/\| z\| $ be the unit tangent at $z$. Then: (i) the direct attributions sum to the lifted angular change,}

\begin{equation}
\begin{aligned}\sum_i\phi_i^{\mathrm{dir}}&=\int_\gamma f^*\omega=\widetilde\theta(1)-\widetilde\theta(0)\\&=\operatorname{wrap}(\hat\theta(x)-\hat\theta(x'))+2\pi k,\end{aligned}
\end{equation}

\emph{with $k$ the integer difference between the lifted and principal wrapped angular changes; (ii) an input coordinate whose Jacobian column $\partial z/\partial x_{i}$ is decoder-null at every point of the path receives exactly zero direct attribution; (iii) the tangent attributions sum to $r r'  \sin (\hat{\theta }(x) -  \hat{\theta }(x' ))$, with $r$ and $r' $ the endpoint output norms, a gauge-dependent quantity in the sense of Proposition~\ref{prop:1}, and the coordinatewise difference between the variants is bounded by a path integral with separate contributions from output-norm scaling and changes in tangent direction (exact identity, bound and proof in Appendix~\ref{sec:N}).}

Direct AIG integrates the closed, non-exact angle form, so it records lifted angular change and annihilates radial motion at every point of the path. Endpoint-projected AIG instead integrates the fixed endpoint covector $\langle r t(z(1)),dz\rangle $, so its sum depends only on the endpoint representations. For $f(a,b)=e^{b}(\sin  a,\cos  a)$ along the path from (0,0) to $(\pi /2,1)$, the endpoint-projected method assigns the scale-only input $b$ approximately $- 2.56$, while direct AIG assigns it zero. On the sampled ResNet18 evaluation paths, the lifted and wrapped endpoint angular changes agree (Appendix~\ref{sec:N}); this does not make direct and endpoint-projected attributions equal. The angle-form construction and its completeness guarantee apply to $S^{1}$. The higher-dimensional baselines labelled angular IG in the results should not be interpreted as this $S^{1}$ construction.

The implemented higher-dimensional and oracle IG constructions are specified in Appendix~\ref{sec:N.4}.

\textbf{Proof of Proposition~\ref{prop:3}.}

(i) By the chain rule,

\begin{equation}
\sum_i (x_i-x_i')\frac{\partial\hat\theta}{\partial x_i}(\gamma(\tau))=\frac{d\widetilde\theta(\tau)}{d\tau}.
\end{equation}

Integration gives the lifted angular change. The angle form is closed on $\mathbb{R} ^{2}\setminus \{0\}$ and becomes exact on the universal cover; the difference between lifted and principal wrapped changes is an integer multiple of $2\pi $.

(ii) The angle form annihilates the radial direction at every nonzero $z$. A decoder-null Jacobian column therefore contributes zero to the direct integrand along the entire path.

(iii) Summing endpoint-projected attributions gives

\begin{equation}
\begin{aligned}\sum_i\phi_i^{\mathrm{tan}}&=\langle r\,t(z(1)),z(1)-z(0)\rangle\\&=rr'\sin(\hat\theta(1)-\hat\theta(0)).\end{aligned}
\end{equation}

Since $\nabla \hat{\theta }(z)=t(z)/\| z\| $, the coordinatewise difference is

\begin{equation}
\begin{aligned}\phi_i^{\mathrm{tan}}-\phi_i^{\mathrm{dir}}&=(x_i-x_i')\int_0^1\\&\!\!\left\langle r\,t(z(1))-\frac{t(z(\tau))}{\|z(\tau)\|},\frac{\partial z}{\partial x_i}(\gamma(\tau))\right\rangle d\tau.\end{aligned}
\end{equation}

Let $\delta (\tau )\in [0,\pi ]$ be the principal angular separation between $z(1)$ and $z(\tau )$. Splitting the covector difference into a scaling term and a tangent-direction term, and using $\| t(z(1))- t(z(\tau ))\| =2\sin (\delta (\tau )/2)$, gives

\begin{equation}
\begin{aligned}|\phi_i^{\mathrm{tan}}-\phi_i^{\mathrm{dir}}|\le{}&|x_i-x_i'|\int_0^1\\&\!\!\left[\left|r-\frac1{\|z(\tau)\|}\right|+\frac{2\sin(\delta(\tau)/2)}{\|z(\tau)\|}\right]\\&\!\!\cdot\left\|\frac{\partial z}{\partial x_i}(\gamma(\tau))\right\|d\tau.\end{aligned}
\end{equation}

The scaling term need not vanish on a constant-prediction path. Such a path can also have nonzero, cancelling coordinate attributions. For example, $f(x_{1},x_{2})=2(\sin (x_{1}- x_{2}),\cos (x_{1}- x_{2}))$ from (0,0) to (1,1) gives direct AIG $(1,- 1)$ and endpoint-projected AIG $(4,- 4)$. \hfill$\square$

Path statistics for the ResNet18 evaluation split of \S\ref{sec:3.3} ($n = 767$, blur baseline, 50 steps): no winding is detected in the sampled prediction paths and the lifted and wrapped changes agree on every image (largest lifted change 5.5 h); the path length exceeds the geodesic displacement by a median factor of 2.0 (8.7 at the 90th percentile); the largest single-step angular change on any path is 0.32 rad, so the lift is unambiguous at 50 steps; and the raw output norm falls below 0.2 on 25 of 767 paths (minimum 0.09), where the atan2 gradient is large but finite.
\subsection{\texorpdfstring{Angular IG on $S^{2}$ and $S^{3}$}{Angular IG on S2 and S3}}\label{sec:N.4}

The direct angle-form construction of \S\ref{sec:3.3} is specific to $S^1$. On the real gaze and quaternion tasks, IG is computed separately for each coordinate of the normalised output $p(x)=\pi(f(x))$ along the straight pixel path from a black baseline, using 50 steps (150 and 200 backward passes, respectively). Let
\begin{equation}
A_i=(x_i-x_i')\int_0^1\partial_i p(x'+\tau(x-x'))\,d\tau.
\end{equation}
At the endpoint, the implementation projects with $P_{p(x)}=I-p(x)p(x)^\top$ and ranks by $\|P_{p(x)}A_i\|_2$. The vector sum completes to the endpoint-tangential component of the chord,
\begin{equation}
\sum_i P_{p(x)}A_i=P_{p(x)}\bigl(p(x)-p(x')\bigr),
\end{equation}
up to numerical integration error. No parallel transport is implemented, and summing attribution magnitudes does not give this vector completeness identity. The $S^2$ directional score 0.096 uses the tangent-CAM direction at pixels ranked by this unsigned IG map, not an IG direction (Appendix~\ref{sec:K}).

The oracle tangent-norm path-integral row in Table~\ref{tab:1} is a third construction: it integrates the unsigned tangent-Jacobian norm $\|P_{p(\tau)}J_i(\tau)\|_2$ at 32 path points. It is distinct from both direct $S^1$ angular IG and endpoint-projected real-task IG; the direct angular completeness claim does not apply to this magnitude integral.

\subsection{\texorpdfstring{Learning can move the representation along $\ker  D\pi $}{Learning can move the representation along ker Dpi}}\label{sec:N.5}

With $z(w) = (w, w -  1)$, $\pi (z) = z/\| z\| $ and $L = \tfrac12 (\pi (z)_{2} -  1)^{2}$, at $w = 1$ the representation is (1, 0) and $g_{z} = (0, - 1)$, yet the update is $\Delta z = \eta (1, 1) + \operatorname{O}(\eta ^{2})$, which has a radial component. A penalty on decoder-null sensitivity also contributes its own parameter gradient and can change predictions and generalisation, which is why \S\ref{sec:8} treats accuracy neutrality as an empirical hypothesis.

\subsection{The 6D decoder comparison}\label{sec:N.6}

\begin{table*}[tp]
\centering
\caption{6D (Gram-Schmidt) pose, five seeds: mean deletion AUC (degrees) for the three-way decoder comparison of \S\ref{sec:7.3}.}
\label{tab:15}
\small
\setlength{\tabcolsep}{4pt}
\begin{tabularx}{\linewidth}{@{}Xccc@{}}
\toprule
Attribution target & Decoder-invariant? & Collapse-free? & Mean AUC (deg) \\
\midrule
Ambient sum over raw $z \in  \mathbb{R} ^{6}$ & no & no & 48.9 \\
Sum over decoded $R$ entries & yes & no & 49.2 \\
Tangent differential through $D\pi $ & yes & yes & 63.0 \\
\bottomrule
\end{tabularx}
\end{table*}

The 6D arm is trained on ResNet18 only, with a Linear$(512,6)$ head. Gram--Schmidt decoding is applied outside the network in both the loss and attribution. CAMs hook the last residual block of layer4 ($512\times7\times7$), as in the quaternion ResNet18 arm. The full-differential map takes the $\ell^2$ norm of three tangent-coordinate Grad-CAM maps with scalar responses $f_i=\langle R,\bar R G_i\rangle_F$, where $\bar R$ is the detached forward rotation and $G_i$ are the three tangent generators. The two scalar target maps are rectified with ReLU; the tangent map is not. The paired comparison with the invariant scalar gives $p=0.0007$ (reported as $p<0.001$).

The rotation-matrix scalar target is decoder-invariant, but it still collapses a multidimensional prediction into one scalar; decoder invariance alone produces almost no change, while retaining the full differential does (\S\ref{sec:7.3}).

\subsection{Reproducibility}\label{sec:N.7}

We will release code, configuration files, synthetic oracle generators, and the aggregated result files behind the tables and figures [GitHub link removed for double blind submission]. Trained checkpoints and the per-image gaze and pose deletion arrays (five seeds each) will be available upon request. The aggregate files support recomputing seed-level tests, the equivalence intervals in Appendix~\ref{sec:O}, the statistics of Table~\ref{tab:14}, and data figures without raw images. Recomputing per-image bootstraps additionally requires the per-image arrays. Recomputing deletion curves, DCS, AMR and accuracy from predictions requires forward passes and the corresponding images. No per-image plant arrays are available. The institutionally held plant data cannot be redistributed; we release its preprocessing and session-grouped splitting procedure.

The oracle generators are self-contained. Gaze uses MPIIGaze with the stated participant split, and pose uses the released PASCAL3D+ grouped split. Attribution results are single-seed unless stated. The plans described as pre-specified (the ten-seed training difference tests, 6D decoder arm, blind-point test, and rectified-tangent cell) were dated comments in the released job-queue script before the corresponding jobs, not an external registration. The equivalence margin was chosen post hoc (Appendix~\ref{sec:O.2}).

\section{Deletion Metric Ablation and Training Controls}\label{sec:O}

\subsection{What the deletion metric rewards}\label{sec:O.1}

Table~\ref{tab:16} shows that deletion rankings and decoder invariance can disagree, since the metric does not always rank attributions for the decoded prediction above attributions for raw coordinates. The scalar a CAM back-propagates is usually left implicit, yet it matters by 1.3 h here, a spread comparable to the differences between CAM variants the literature treats as meaningful. The two targets that satisfy the invariance condition of Proposition~\ref{prop:1}, tangent projection and the decoded angle, agree to 0.012 h (the positive per-image scaling preserves rankings except for numerical ties; details below) and both score lowest, while the four that do not span 0.965 h and cos-only scores highest. The $\sin  + \cos $ row is exactly the ambient Grad-CAM baseline used throughout \S\ref{sec:7}, so its 2.571 h uses the same target definition, evaluated here on $n=100$ rather than the $n=767$ main comparison.

In the implementation, the decoded-angle map is the tangent map scaled by $1/(\|z\|^2+10^{-8})$ through GAP weighting, ReLU, absolute value and upsampling. Rectification leaves exactly zero attribution on 6--50\% of pixels. Once deletion exceeds the nonzero fraction, top-$k$ selection can break zero-attribution ties differently between arrays. Float32 rounding also swaps 2--6 nearly tied pixels at fractions 0.2--0.7 without changing the reported prediction displacement. The per-fraction means agree at 0.1 and differ only at 0.8 (2.997 versus 2.972 h) and 0.9 (4.382 versus 4.298 h), explaining the residual mean-AUC difference. Standard-IG target selection is detailed in Appendix~\ref{sec:A}.

\begin{table}[tp]
\centering
\caption{Grad-CAM scalar target target, plant $S^{1}$, ResNet18, $n = 100$, deletion AUC ($h$, higher means greater deletion displacement) under the normalised metric.}
\label{tab:16}
\small
\setlength{\tabcolsep}{4pt}
\begin{tabularx}{\linewidth}{@{}Xcc@{}}
\toprule
Target & Deletion AUC (h) & std \\
\midrule
cos only & 2.946 & 1.514 \\
Output norm $\| y\| $ & 2.654 & 2.195 \\
$\sin  + \cos $ & 2.571 & 1.663 \\
sin only & 1.981 & 1.618 \\
Decoded angle & 1.655 & 1.205 \\
Tangent projection & 1.643 & 1.200 \\
\bottomrule
\end{tabularx}
\end{table}

We do not read this as showing that the deletion metric is wrong. Deletion measures the change in the decoded prediction after a finite input perturbation, and a map built for an ambient coordinate can still select pixels that move the decoded prediction, particularly when scale and angle depend on shared features. A local gradient map can also score lower on finite deletion because of saturation, nonlinear effects, feature interactions or masking artifacts. What the table shows is a disagreement between an attribution's invariance property and its finite-perturbation ranking; establishing that the metric itself fails would require independent ground truth, which only the oracles provide.

\subsection{Training controls (Section~\ref{sec:8})}\label{sec:O.2}

\begin{table*}[tp]
\centering
\caption{Radial-penalty controls, ResNet18, plant $S^{1}$, 10 epochs, normalised angular metric. Val err: mean over 3 initialisation seeds; 10-seed: ten fresh seeds per arm with pre-specified tests; $\rho _{\mathrm{rad}}$: radial gradient fraction.}
\label{tab:17}
\small
\setlength{\tabcolsep}{4pt}
\begin{tabularx}{\linewidth}{@{}lXccc@{}}
\toprule
Arm & Description & Val err (h) & 10-seed & $\rho _{\mathrm{rad}}$ \\
\midrule
A & Baseline (tanh head, no penalty) & 1.076 & 1.021 & 0.200 \\
$B$ & $L^{2}$-normalized head & 1.069 & $n/a$ & $7\times10^{-15}$ \\
$C$ & Radial penalty, $\lambda =0.1$ & 1.087 & 1.019 & 0.167 \\
$D$ & Radial penalty, $\lambda =1.0$ & 1.780 & $n/a$ & 0.143 \\
$E$ & Generic grad-norm penalty & 1.083 & $n/a$ & 0.194 \\
$F$ & Tangential penalty, $\lambda =0.1$ & 1.131 & 1.025 & 0.261 \\
$G$ & Weight decay $10^{-4}$ & 1.111 & $n/a$ & 0.203 \\
\bottomrule
\end{tabularx}
\end{table*}

The strong-penalty arm $D$ has higher rescored error (1.780 versus 1.076 h), but its checkpoints were selected under the earlier unnormalised metric, at epoch 1 in all three seeds. This combines penalty and checkpoint-selection effects. The weak radial penalty changes the measured radial gradient fraction from 0.200 to 0.167, a reduction of 0.033 rather than elimination of radial sensitivity.

In the ten-seed ResNet18 plant comparison (seed-paired, df=9), $F-A$ is $+0.0036$ h (two-sided $p=0.54$), $F-C$ is $+0.0056$ h ($p=0.49$), and $C-A$ is $-0.0021$ h ($p=0.79$). The respective 90\% intervals are $[-0.0067,+0.0139]$, $[-0.0086,+0.0198]$, and $[-0.0157,+0.0116]$ h. All lie strictly within $\pm0.02$ h, but that equivalence margin was chosen after inspecting the data: TOST was added on 5 November, whereas the pre-specified plan named only the three difference tests. The planned $F>A$ and $F>C$ tests were one-sided ($p=0.27$ and 0.24). Thus the equivalence result is exploratory; these intervals do not establish equivalence within $\pm0.01$ h.

Separately, the pose control with an unnormalised quaternion head has a reported error difference of $- 0.3\pm 1.0{}^\circ $ relative to the normalised head, with no detected difference between radial and tangential penalties. This degree-valued pose comparison does not establish the plant task's hour-valued equivalence margin.

A common circular-regression implementation computes angular error as arccos of the unnormalised dot product with the unit-norm target, which evaluates to $\arccos (\operatorname{clamp}(\| \hat{y}\|  \cos  \Delta \theta , - 1, 1))$ and mixes prediction norm into the angle. For angular errors below $\pi /2$, norms below 1 inflate the reported error and norms above 1 reduce it, with clipping to zero once $\| \hat{y}\| \cos  \Delta \theta \ge 1$. The direction reverses for angular errors above $\pi /2$. In both regimes the metric depends on representation norm rather than only on the decoded angle. Any intervention that moves $\| \hat{y}\| $ therefore changes the reported error and the selected epoch without changing the decoded prediction, and the penalties studied here do move it: the mean prediction norm is 0.96 for the baseline, 1.24 under the strong radial penalty and 0.92 under the tangential penalty. All angular errors in this paper use the normalised metric. The ten-seed arms were also selected under it; the three-seed arms of Table~\ref{tab:17} and Appendix~\ref{sec:I}, and the four $S^{1}$ backbone checkpoints of \S\ref{sec:4}, were selected under the unnormalised metric and re-scored with the normalised one, which is the selection caveat noted for arm $D$ above. The $L^{2}$-normalised output head (arm $B$), standard in gaze and pose estimation, enforces $\| \hat{y}\|  = 1$ directly and has a similar mean error to the baseline (1.069 h vs 1.076 h, three seeds).

\subsection{Decoder-null sensitivity regularisation}\label{sec:3.4}\label{sec:O.3}

For any loss of the form $L(\pi (z))$, $\partial L/\partial z = D\pi (z)^{\top }\nabla L \in  (\ker  D\pi (z))^{\perp}$: the loss gradient with respect to the representation has no decoder-null component. This does not confine learning to the tangent space, because a gradient step changes the representation by $\Delta z = - \eta  J_{w} J_{w}^{\top } g_{z} + \operatorname{O}(\eta ^{2})$ and $J_{w} J_{w}^{\top }$ need not preserve the split (Appendix~\ref{sec:N}), and the identity excludes the ambient MSE that trains the oracles and the real-task $S^{1}$ models. Whether suppressing decoder-null sensitivity affects accuracy is therefore an empirical question (\S\ref{sec:8}). The general penalty is $\| P_{\ker  D\pi } J_{f}\| ^{2}$, with $J_{f}$ the input Jacobian; we implement the scale-invariant version

\begin{equation}
L_{\mathrm{null}}=\lambda\,\mathbb E\!\left[\frac{\|P_{\ker D\pi}J_f\|^2}{\|J_f\|^2}\right],
\end{equation}

which on $S^{1}$ is the radial fraction of the output gradient's squared norm. The normalisation is necessary because the unnormalised penalty is minimised by shrinking \emph{all} sensitivity, close to a generic gradient-norm penalty, which is one of our controls (arm $E$, \S\ref{sec:8}). An $L^{2}$-normalised head drives the fraction to zero by construction, and a penalty on the \emph{tangential} fraction serves as a direction control. If the penalty is zero at $x$, every column of $J_{f}(x)$ is horizontal and $P_{T} J_{f}(x)=J_{f}(x)$. This removes radial contamination but does not make a fixed scalar readout equivalent to the full differential, and it does not establish equality of CAM maps at internal layers.

\subsection{Decoder-Null Sensitivity and Learning}\label{sec:8}\label{sec:O.4}

The decoded-loss gradient identity in \S\ref{sec:3.4} does not apply to the ambient MSE used to train the plant models. The experiments here are therefore a separate empirical study of decoder-null regularisation, rather than a test of accuracy neutrality implied by that identity. We compare seven arms with three seeds and weak-penalty arms with ten fresh seeds on ResNet18 plant $S^{1}$ (Table~\ref{tab:17}; Figure~\ref{fig:9}). The weak radial penalty changes the measured radial fraction from 0.200 to 0.167 in the three-seed controls, so the null accuracy result concerns a modest reduction rather than elimination of radial sensitivity.

For the ten-seed contrasts, paired two-sided tests give $p=0.54$, 0.49, and 0.79. The 90\% intervals lie strictly within $\pm0.02$ h, including the $F-C$ upper endpoint $+0.0198$ h, but this margin was selected post hoc. The equivalence analysis is therefore exploratory; the planned one-sided $F>A$ and $F>C$ tests give $p=0.27$ and 0.24 (Appendix~\ref{sec:O.2}). The strong-penalty arm has higher error (1.780 versus 1.076 h), but its checkpoints were selected under the earlier norm-dependent metric, combining penalty and selection effects. Cross-backbone plant controls are in Appendix~\ref{sec:I} and pose controls in Appendix~\ref{sec:O.2}. All errors reported here are rescored with the normalised angular metric.

\section{Additional Real-Task Results}\label{sec:P}

\subsection{Gaze: uncertainty and additional comparisons}\label{sec:P.1}

Five ResNet18 gaze models were trained from independent initialisations on the identical leave-one-person-out split; attribution uses validation participant p13 ($n=2,996$), also used for epoch selection (Table~\ref{tab:2}). Tangent Grad-CAM's advantage over ambient Grad-CAM is concentrated at low deletion fractions, where it is large and reliable: at the 10\% fraction tangent reaches $6.50 \pm  0.29{}^\circ $ against $4.52 \pm  0.36{}^\circ $, 44\% larger deletion displacement, favouring tangent in $5/5$ seeds ($p = 0.001$, paired over seeds). The advantage holds at $p < 0.05$ through fraction 0.5 and reverses from 0.8 onward, so the full-range mean ($+13.5$\% $\pm  7.8$\%, $p = 0.017$) obscures the reversal at high deletion fractions (Figure~\ref{fig:5}, Appendix~\ref{sec:A}). Training variability rather than image sampling is the binding uncertainty: the seed-level 95\% interval at 10\% is about six times wider than the per-image bootstrap (Appendix~\ref{sec:B}), and the earlier single-checkpoint improvement in full-range mean deletion AUC (23\%) exceeds the corresponding per-seed mean-AUC improvements (5.2\% to 22.5\%). These percentages use the full range, unlike the 44\% gain at fraction 0.1.

This conclusion is limited to the seeded Grad-CAM comparison. The seeded comparison covers tangent and ambient Grad-CAM only; the perturbation baselines (Table~\ref{tab:5}, Appendix~\ref{sec:A}) ran on a single checkpoint, where the original spherical-occlusion estimator leads at fractions 0.1--0.3. The centred estimator (patch 32, stride 16, 169 passes, footprint overlap-add) instead leads tangent Grad-CAM at every fraction: its mean is $11.02^\circ$, versus $9.37^\circ$ for the original estimator. Both use the same checkpoint, 2,996 p13 images, black fill and unsigned geodesic displacement; grid resolution and patch-to-pixel mapping differ. Appendix~\ref{sec:A} reports all nine centred values. We claim the like-for-like result only: swapping the ambient scalar target for the tangent projection, changing nothing else, gives a reliable early-fraction gain on one backbone and one dataset.

The mechanism is the collapse of \S\ref{sec:6.4} rather than radial removal: the gaze head is hard-normalised and its radial gradient fraction is zero at machine precision, so the sum scalar target reads the two-dimensional tangent differential through one fixed slice and the tangent magnitude does not. The dimension count $(n- m)/m$ does not explain the pattern, since it is lower on $S^{2} (0.5)$ than on $S^{1} (1.0)$ while the gain is larger on $S^{2}$.

The endpoint-projected higher-dimensional IG baseline (Appendix~\ref{sec:N.4}) scores $7.68^\circ$ mean deletion displacement on $S^2$. Its map is unsigned: the evaluator ranks pixels by IG magnitude but takes directions from the tangent-CAM field. The resulting cosine 0.096 is therefore a tangent-CAM direction proxy at IG-ranked pixels, not an IG direction. Tangent CAM's own ranking gives cosine 0.066. Both use approximately 300,000 pixel deletions. Appendix~\ref{sec:K} illustrates a rescaling of the directional metrics under an assumed error model; it does not establish an IG-to-IG directional comparison.

\begin{table*}[tp]
\centering
\caption{MPIIGaze $S^{2}$ deletion (degrees, higher means greater deletion displacement), tangent vs ambient-sum Grad-CAM, five independent initialisations (ResNet18, validation participant p13, $n=2,996$ per run, also used for epoch selection). Seed columns: mean $\pm $ SD over runs, uncorrected paired $t$-test (df $= 4$); bootstrap: paired per-image differences averaged over seeds. Fractions are nested, so tests are descriptive. Full version in Table~\ref{tab:8}.}
\label{tab:2}
\small
\setlength{\tabcolsep}{4pt}
\begin{tabularx}{\linewidth}{@{}Xccccc@{}}
\toprule
 & Deletion (deg) &  & Seed-level $(n=5)$ &  & Bootstrap \\
\midrule
Fraction & Tangent & Ambient-sum & $\Delta  \pm $ SD & $p$ & $\Delta $ [95\% CI] \\
\midrule
0.1 & $6.50\pm 0.29$ & $4.52\pm 0.36$ & $+1.97\pm 0.47$ & \textbf{0.001} & $+1.97 [+1.88, +2.07]$ \\
0.4 & $9.15\pm 0.69$ & $7.43\pm 0.59$ & $+1.72\pm 0.58$ & \textbf{0.003} & $+1.72 [+1.59, +1.86]$ \\
0.7 & $9.88\pm 1.23$ & $9.81\pm 1.00$ & $+0.06\pm 0.83$ & 0.870 & $+0.06 [-0.01, +0.14]$ \\
0.9 & $10.16\pm 1.24$ & $10.51\pm 1.29$ & $-0.35\pm 0.88$ & 0.416 & $-0.35 [-0.39, -0.32]$ \\
Mean & $9.08\pm 0.72$ & $8.02\pm 0.65$ & $+1.06\pm 0.60$ & \textbf{0.017} & $n/a$ \\
\bottomrule
\end{tabularx}
\end{table*}

\subsection{Plant: method, backbone, and baseline comparisons}\label{sec:P.2}

Table 5 reports plant attribution on one checkpoint per backbone, using the validation split. Ambient-sum LayerCAM and ambient HiResCAM use the same scalar target but differ in gradient rectification. Their deletion scores are 5.06 versus 2.11 h on ResNet18, 4.45 versus 0.94 h on ResNet50, and 3.53 versus 0.65 h on EfficientNet; their ViT scores are nearly equal. The two-polarity tangent LayerCAM variant scores $5.24/4.53/4.33/1.44 h$, compared with $5.06/4.45/3.53/0.75 h$ for standard ambient-sum LayerCAM. Its numerical gains are $0.18/0.08/0.80/0.69 h$. This comparison changes both the target and polarity aggregation and does not establish a seed-replicated advantage. The unrectified tangent Grad-CAM scores are $1.58/1.32/1.94/0.73 h$, compared with $2.10/0.94/0.65/1.10 h$ for the ambient sum.

CAM scores are lower on the ViT than input-space occlusion and IG scores, which exceed 8 h. The head reads the CLS token, while the hooked layer is the first LayerNorm in the final block; the result is therefore specific to this layer choice. Occlusion is the strongest perturbation method on $S^{1}$ and requires 729 forward passes per image. Wrapping its angular difference changes deletion AUC by about 0.01 h or less. The signed geodesic accumulation gives lower scores than the unsigned version, consistent with cancellation between opposite-signed displacements from overlapping patches. Circular LIME has higher scores than scalar-angle LIME on all four checkpoints; Appendix~\ref{sec:F} gives a separate kernel/surrogate ablation.

The IG family has the highest scores among the gradient methods. The angular variants reach approximately 8.40 h, and direct angular IG differs from the best standard scalar-target IG by at most 0.033 h in Table~\ref{tab:6}. Direct AIG is complete for the lifted angular change. Its blur-baseline DCS estimates are $0.599/0.534/0.524/0.735$ across the four backbones. The DCS estimates of 0.52--0.53 are not distinguishable from chance under approximate Wilson intervals that use the number of contributing images as the effective sample size. These intervals approximate dependence among pixel deletions; they do not directly estimate between-image variability or provide calibrated clustered uncertainty (Appendix~\ref{sec:Q}). Black-baseline results cover four backbones; the distributional-baseline evaluation covers ResNet18 only. Single-point tangent-CAM direction is also checkpoint-dependent: ResNet18 scores 0.661 in the headline run and 0.487 in the $\lambda =0$ retrain. These results support reporting directional performance separately from unsigned deletion and completeness.

The direct and endpoint-projected $S^1$ IG rows in Figure~\ref{fig:8} and Table~\ref{tab:7} are distinct; the black and distributional direct-AIG results are reported separately in Appendix~\ref{sec:E}.

\subsection{Pose: statistical and architecture comparisons}\label{sec:P.3}

PASCAL3D+ provides a real rotation task with five initialisation seeds per arm. With the 6D Gram-Schmidt decoder, we compare the sum of six raw coordinates, the sum of nine decoded rotation-matrix entries, and the full differential through $D\pi $. Mean deletion AUC is $63.0{}^\circ $ for the full differential, $48.9{}^\circ $ for the ambient sum (difference $+14.1{}^\circ $, $p=0.008$), and $49.2{}^\circ $ for the decoder-invariant scalar (difference $+13.8{}^\circ $, $p=0.0007$). No difference is detected between the scalar targets ($p=0.85$; Table~\ref{tab:15}). This comparison shows that decoder invariance of one scalar target does not match the full differential's deletion score in this setting. Proposition~\ref{prop:2} does not establish optimality of either scalar for this decoder.

For the ResNet18 quaternion head (Table~\ref{tab:4}), tangent Grad-CAM exceeds the ambient sum in all five seeds at deletion fractions 0.1--0.7, with paired seed-level $p<0.05$ over that range. The mean difference across fractions is $+8.77{}^\circ  (p=0.013)$. The per-image bootstrap over 4,886 validation crops excludes zero at all nine fractions (Figure~\ref{fig:4}, left). The gap is associated with low alignment, although alignment also covaries with pose difficulty (\S\ref{sec:6.4}). HiResCAM coincides with Grad-CAM under the GAP structure used here (Appendix~\ref{sec:H}).

\begin{table*}[tp]
\centering
\caption{PASCAL3D+ $\operatorname{SO}(3)$ deletion (degrees of geodesic distance, higher means greater deletion displacement), tangent vs ambient-sum Grad-CAM, five independent initialisations (ResNet18, $n = 4,886$ per run). Seed columns report mean $\pm $ SD with an uncorrected paired $t$-test (df $= 4$); the bootstrap column resamples paired per-image differences.}
\label{tab:4}
\small
\setlength{\tabcolsep}{4pt}
\begin{tabularx}{\linewidth}{@{}Xccccc@{}}
\toprule
Fraction & Tangent & Ambient-sum & $\Delta  \pm $ SD & $p$ & $\Delta $ [95\% CI] \\
\midrule
0.1 & $26.42\pm 3.13$ & $21.68\pm 3.88$ & $+4.74\pm 2.59$ & 0.015 & $+4.74 [+4.39, +5.10]$ \\
0.4 & $56.92\pm 4.19$ & $44.90\pm 5.92$ & $+12.02\pm 7.56$ & 0.024 & $+12.02 [+11.30, +12.71]$ \\
0.7 & $67.40\pm 6.91$ & $58.47\pm 4.80$ & $+8.93\pm 5.96$ & 0.029 & $+8.93 [+8.19, +9.66]$ \\
Mean & $56.55\pm 4.82$ & $47.77\pm 3.48$ & $+8.77\pm 4.65$ & 0.013 & $+8.77 [+8.31, +9.24]$ \\
\bottomrule
\end{tabularx}
\end{table*}

Ambient-sum LayerCAM exceeds the plain tangent map through fraction 0.6 (mean tangent-minus-ambient difference $- 2.55{}^\circ $, $p=0.021$). For the two-polarity tangent LayerCAM variant, no difference from ambient-sum LayerCAM is detected (mean difference $- 0.08{}^\circ $, $p=0.75$; 90\% CI $[- 0.58{}^\circ , +0.42{}^\circ ]$). The two-polarity variant exceeds the plain tangent map by $2.47{}^\circ  (p=0.008)$. Because it changes rectification and polarity aggregation, this result does not isolate a target-only effect. On this hard-normalised head the radial derivative is already zero, so the distinction concerns scalar direction loss and map aggregation. The two-polarity variant is not invariant to arbitrary rotations of its tangent basis (\S\ref{sec:3.1}).

ResNet50 gives a positive mean quaternion gain, but one of five seeds favours the ambient sum and the seed-level test is not significant at 0.05. Across five seeds tangent Grad-CAM leads the ambient sum by $+7.54{}^\circ $ on average ($p = 0.094$, df $= 4$): four seeds favour the tangent map by $6{}^\circ $ to $19{}^\circ $, one reverses it by $2{}^\circ $, and the paired per-image bootstrap excludes zero at all nine fractions (pooled $+7.54{}^\circ $, 95\% CI $[+7.22{}^\circ , +7.86{}^\circ ]$). Ambient-sum LayerCAM again leads the plain tangent map $(- 2.97{}^\circ , p = 0.039)$. On ViT-$B/16$ the picture changes. Across five seeds, the same unrectified tangent map (tangent LayerCAM equals tangent HiResCAM on the quaternion head) exceeds ambient HiResCAM by $+5.63^\circ$ ($p=0.043$) and ambient-sum LayerCAM at $p=0.047$. The hook is the first LayerNorm of the final encoder block. The two-polarity variant was not run on ViT. The tangent map loses the GAP-weighted comparison, consistent with Appendix~\ref{sec:H}: spatial averaging becomes a poor approximation once the architecture no longer has the GAP structure that motivates it.

Perturbation methods put this in context. On a single quaternion checkpoint, occlusion leads at low fractions ($47.5{}^\circ $ at 0.1 against $28.1{}^\circ $ for the tangent map) and on the mean $(57.2{}^\circ )$, with angular IG close behind, at 169 forward or 200 backward passes per image against four for the tangent map. The results support gains for the specified full-differential comparisons, rather than a uniform advantage over every CAM baseline, architecture, or seed.

\section{Implementation and Evaluation Details}\label{sec:Q}

Evaluation covers localisation, direction, and attribution completeness. Synthetic oracles have analytically known relevant and irrelevant regions. Directional metrics assess the sign or direction of prediction changes. Completeness compares attribution sums with a specified target change: the lifted angle for direct angular IG, or the chosen coordinate for ambient IG. Real-task finite-deletion displacement rankings need not agree with decoder invariance (Appendix~\ref{sec:O}).

\textbf{Metrics.} \emph{Oracle AMR: on the two-stream models, the fraction of absolute attribution mass in the region that provably encodes the angle (1.0 perfect, 0.5 chance),} the only metric with ground truth. \emph{Geodesic deletion AUC}: rank pixels by $|\phi _{i}|$, set the top 10\%, \ldots{}, 90\% to zero in all channels in raw $[0,1]$ space, measure the geodesic displacement of the decoded prediction and average over the nine fractions (hours on $S^{1}$, degrees on $S^{2}$ and $\operatorname{SO}(3)$; higher means greater deletion displacement). \emph{Directional consistency (DCS)} on $S^{1}$: delete the top-100 positively attributed pixels individually and record the fraction that move the prediction against the attributed direction (chance 0.5). \emph{Directional cosine} on $S^{2}$: mean cosine between the predicted tangent direction and the negative actual tangential displacement. Insertion does not discriminate between methods on $S^{1}$ (Appendix~\ref{sec:G}).

For each selected pixel $i$, let $x^{\setminus i}$ set all its RGB channels to zero and let $\Delta_i=\operatorname{wrap}_{[-12,12)}(\hat\theta(x^{\setminus i})-\hat\theta(x))$ in hours. Pooling the selected positive-attribution pixels $\mathcal P$ across images gives
\begin{equation}
\mathrm{DCS}=\frac{1}{|\mathcal P|}\sum_{(x,i)\in\mathcal P}\mathbf1\{\phi_i\Delta_i<0\}.
\end{equation}
A zero change is a failure. There are 76,700 deletions for CAMs ($767\times100$) and 74,300 for angular IG ($743\times100$): 24 images with no positive IG pixel are skipped. The reported image-effective Wilson intervals use 767 and 743 contributing images, respectively. This effective sample-size approximation does not turn the pooled pixel score into an unweighted image-level estimate. On gaze, the IG-labelled directional cosine uses the tangent-CAM direction at IG-ranked pixels, as described in Appendix~\ref{sec:K}.

\textbf{Synthetic oracles.} $64\times 64$ two-column images: the left half encodes the angular target, the right half an irrelevant magnitude factor as a grating whose contrast sets the magnitude. A two-stream network consumes them through disjoint branches, so ground-truth attribution is exactly the left half and the guarantee is architectural. The rotation oracle predicts a unit quaternion, so its manifold is $S^{3} \subset  \mathbb{R} ^{4}$ (Appendix~\ref{sec:M}). The Table~\ref{tab:1} test sets have $n=500/30/200$ for $S^{1}/S^{2}/S^{3}$; Appendix~\ref{sec:I.2} describes the separately trained wider-range, stratified $n=2,000$ blind-point analysis. The \emph{entangled} single-stream controls on $S^{1}$ and $S^{2}$ have $n = 200$. Details in Appendix~\ref{sec:L}.

\textbf{Plant circadian $S^{1}$.} 3,765 segmentation-masked images, timestamps from filenames, session-grouped split (train 2,621 / val 767 / test 377). Angular error is in hours (h). Four backbones share a $\operatorname{Linear}(d,2)+\tanh $ head and reach 1.30--1.48 h test error against 6.00 h random and 2.32 h constant baselines (Appendix~\ref{sec:C}; checkpoint selection in Appendix~\ref{sec:O}). Attribution uses the 767-image validation split, which was also used for checkpoint selection; the 377-image test split is used for the accuracy values reported here.

\textbf{Gaze $S^2$.} MPIIGaze~\cite{zhang2015}, ResNet18, one LOPO fold (p14 test, p13 validation; 15 epochs, Adam, $10^{-4}$, batch 128). Accuracy is reported on p14 ($n=3,000$): the original unseeded checkpoint has test mean $5.20^\circ$ and p13 validation error $5.49^\circ$. All gaze attribution metrics use p13 ($n=2,996$), which is also the epoch-selection participant. The five seeded attribution models have validation errors $5.63/5.29/5.36/5.38/5.23^\circ$ and no test evaluation; Appendix~\ref{sec:C} distinguishes these from the original checkpoint.

\textbf{Pose $\operatorname{SO}(3)$.} PASCAL3D+ \cite{xiang2014}: 24,308 object crops from the nine categories without rotational symmetry, split $70/20/10$ grouped by source photograph (seed 42), validation $n = 4,886$. ResNet18, ResNet50 and ViT-$B/16$ with a quaternion head (Linear to $\mathbb{R} ^{4}$, $L^{2}$ normalisation), and a ResNet18-only 6D head decoded by Gram-Schmidt \cite{zhou2019}; 20 epochs, Adam, $10^{-4}$, batch 64, five seeds. Median validation geodesic error $15.6{}^\circ $ (quaternion, ResNet18), $14.4{}^\circ $ (ResNet50), $13.6{}^\circ  (6D)$, $30.2{}^\circ $ (ViT), with the double-cover geodesic $2 \arccos |\langle q, q' \rangle |$. We evaluate rotation only (3-DoF).

The 6D head is Linear$(512,6)$, with Gram--Schmidt decoding outside the network in the loss and attribution. Its CAM hook is the last residual block of layer4 ($512\times7\times7$), as for quaternion ResNet18. The full map is the $\ell^2$ norm of three tangent-coordinate Grad-CAM maps, using the detached forward frame; the two scalar maps use ReLU and the tangent map does not (Appendix~\ref{sec:N.6}). The ViT quaternion hook is the first LayerNorm of the final encoder block; its per-location tangent comparison is unrectified, not the two-polarity variant.

\end{document}